\documentclass[10pt,twocolumn,letterpaper]{article}

\usepackage{cvpr}              

\usepackage{times}
\usepackage{epsfig}

\usepackage{graphicx}
\usepackage{amsmath}
\usepackage{amssymb}
\usepackage{booktabs}
\usepackage{physics}
\usepackage{makecell}
\DeclareMathSymbol{\shortminus}{\mathbin}{AMSa}{"39}
\newcommand*\diff{\mathop{}\!\mathrm{d}}
\usepackage{ifthen}
\newboolean{supponly}

\usepackage[pagebackref,breaklinks,colorlinks]{hyperref}

\setboolean{supponly}{false}  

\def\paperID{65} 
\def\confName{3DV\xspace}
\def\confYear{2024\xspace}

\usepackage{color}
\usepackage{colortbl}
\definecolor{turquoise}{cmyk}{0.65,0,0.1,0.3}
\definecolor{purple}{rgb}{0.65,0,0.65}
\definecolor{dark_green}{rgb}{0, 0.5, 0}
\definecolor{orange}{rgb}{0.8, 0.6, 0.2}
\definecolor{red}{rgb}{0.8, 0.2, 0.2}
\definecolor{darkred}{rgb}{0.6, 0.1, 0.05}
\definecolor{blueish}{rgb}{0.0, 0.3, .6}
\definecolor{light_gray}{rgb}{0.7, 0.7, .7}
\definecolor{pink}{rgb}{1, 0, 1}
\definecolor{greyblue}{rgb}{0.25, 0.25, 1}
\definecolor{gold}{rgb}{0.7, 0.5, 0}
\definecolor{Red}{rgb}{1.0, 0.7, 0.7}
\definecolor{Orange}{rgb}{1.0, 0.85, 0.7}
\definecolor{Green}{rgb}{0.7, 1.0, 0.7}

\newcommand{\methodname}{MELON\xspace}
\newcommand{\titlemethod}{MELON: NeRF with Unposed Images in SO(3)\xspace}
\newcommand{\lossname}{modulo loss\xspace}

\newcommand{\LossName}{Modulo Loss\xspace}

\newcommand{\position}{\mathbf{r}}

\newcommand{\viewdir}{\mathbf{d}}
\newcommand{\density}{\sigma}
\newcommand{\radiance}{\mathbf{c}}

\newcommand{\latent}{z_i}

\newcommand{\nerfcolor}{C}

\newcommand{\image}{I_i}
\newcommand{\pixel}{\mathbf{p}}

\newcommand{\pose}{R_i}

\newcommand{\sotwo}{\text{SO}(2)}

\newcommand{\shared}{f}
\newcommand{\torus}{f}
\newcommand{\ang}{\theta_i}

\newcommand{\pix}{\theta}
\newcommand{\plane}{\mathcal{P}}
\newcommand{\segment}{S}
\newcommand{\loss}{\mathcal{L}}
\newcommand{\enc}{h}
\newcommand{\weightsenc}{\xi}
\newcommand{\weightsshared}{\psi}
\newcommand{\degen}{\mathcal{S}}
\newcommand{\latentspace}{\Omega}

\newcommand{\relation}{\mathcal{R}}

\newcommand{\alignment}{z_0}
\newcommand{\reporder}{N}
\newcommand{\azim}{\theta}
\newcommand{\elev}{\phi}
\newcommand{\conv}{K}
\newcommand{\roll}{\alpha}
\newcommand{\sothree}{\text{SO}(3)}
\newcommand{\sethree}{\text{SE}(3)}

\begin{document}

\title{
\titlemethod
}

\author{Axel Levy\footnote[1]{denotes equal contribution}\\
Stanford University\\
{\tt\small axlevy@stanford.edu}
\and
Mark Matthews\footnote[1]{denotes equal contribution}\\
Google\\
{\tt\small mjmatthews@google.com}
\and
Matan Sela \\
Google\\
{\tt\small matansel@google.com}
\and 
Gordon Wetzstein \\
Stanford University\\
{\tt\small gordon.wetzstein@stanford.edu}
\and
Dmitry Lagun\\
Google\\
{\tt\small dlagun@google.com}
\and
\vspace{-5mm}
\phantom{XXXXXXXXXXXXXXXXXXXXXXXXXXXXXXXXXXXXXXX}\\
\vspace{-5mm}
{\small \url{https://melon-nerf.github.io}}
}

\ifthenelse{\boolean{supponly}}{

\appendix

\setcounter{page}{1}

\renewcommand{\thefigure}{S\arabic{figure}}
\renewcommand{\thetable}{S\arabic{table}}
\setcounter{figure}{0}  
\setcounter{table}{0}  

\twocolumn[
\centering
\Large
\textbf{\titlemethod} \\
\vspace{0.5em}Supplementary Material \\
\vspace{1.0em}
] 

\section{Datasets}

\subsection{1D Dataset}

Fig.~\ref{fig:1d-dataset} shows an example of a ground truth 1D function $f^*$, its self-similarity maps and its regions of attraction. We show visual results for the ablation study of Fig.~5. Details on the generation process are given in the main paper (4.1). A jupyter notebook for generating these datasets and running our experiments will be provided in an open source repository.

\begin{figure}[h]
    \centering
    \includegraphics[width=\linewidth]{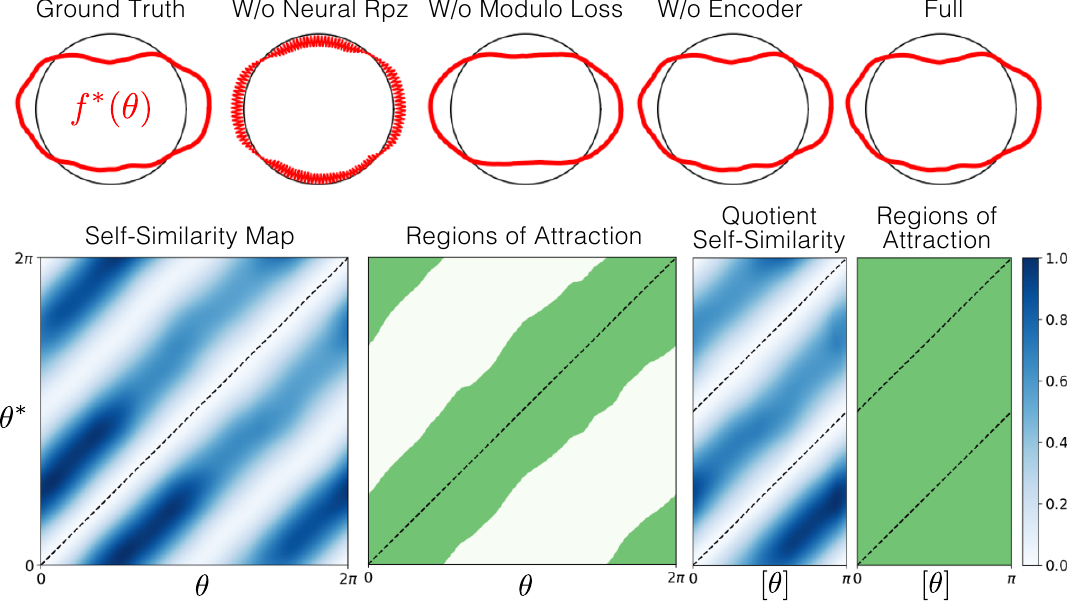}
    \caption{(Top) Example of ground truth function $f^*$ and visual results for the ablation study (Fig.~5).
    (Bottom) Self-similarity maps and regions of attraction. The self-similarity map shows local minima on the line $\theta^*=\theta+\pi$. With the equivalence relation $\relation=\relation_2$, the regions of attraction cover the entire quotient set $\Omega/\relation$.
    }
    \label{fig:1d-dataset}
\end{figure}

\subsection{3D Datasets}

Table~\ref{tab:datasets} summarizes the properties of the datasets used for 3D inverse rendering. Fig.~\ref{fig:sph-dataset} shows sample views of the RGB-MELON datasets. These datasets and the script for generating them in Blender \cite{blender} will be provided upon publication.

\begin{table}[ht]
  \centering
  \resizebox{\linewidth}{!}{
  \begin{tabular}{l|cccccccc}
  \toprule
  
  Dataset &
  Synth. &
  Res. &
  Train &
  Test &
  Roll $=0$ &
  Elev. Range &
  $\reporder$ &
  Encoder Res. \\
  
  \midrule
  
  RGB-M &
  Yes &
  128 &
  100 &
  16 &
  Yes &
  $\{0^\circ\}$ &
  1-4  &
  $64\times64$ \\
  
  \midrule

  Lego~\cite{Mildenhall20eccv_nerf} &
  Yes &
  128-400 &
  2-100 &
  200 &
  Yes &
  $[0^\circ,90^\circ]$ &
  2 &
  $32\times32$ \\
  
  Hotdog~\cite{Mildenhall20eccv_nerf} &
  Yes &
  400 &
  100 &
  200 &
  Yes &
  $[0^\circ,90^\circ]$ &
  2 &
  $128\times128$ \\

  Chair~\cite{Mildenhall20eccv_nerf} &
  Yes &
  400 &
  100 &
  200 &
  Yes &
  $[0^\circ,90^\circ]$ &
  2 &
  $128\times128$ \\

  Drums~\cite{Mildenhall20eccv_nerf} &
  Yes &
  400 &
  100 &
  200 &
  Yes &
  $[0^\circ,90^\circ]$ &
  2 &
  $32\times32$ \\

  Mic~\cite{Mildenhall20eccv_nerf} &
  Yes &
  400 &
  100 &
  200 &
  Yes &
  $[0^\circ,90^\circ]$ &
  2 &
  $32\times32$ \\

  Ship~\cite{Mildenhall20eccv_nerf} &
  Yes &
  400 &
  100 &
  200 &
  Yes &
  $[0^\circ,90^\circ]$ &
  2 &
  $64\times64$ \\

  Materials~\cite{Mildenhall20eccv_nerf} &
  Yes &
  400 &
  100 &
  200 &
  Yes &
  $[-90^\circ,90^\circ]$ &
  4 &
  $64\times64$ \\

  Ficus~\cite{Mildenhall20eccv_nerf} &
  Yes &
  400 &
  100 &
  200 &
  Yes &
  $[-90^\circ,90^\circ]$ &
  2 &
  $128\times128$ \\

  \midrule
  
  Cape~\cite{goldcape} &
  No &
  400 &
  99 &
  2 &
  No &
  $[0^\circ,90^\circ]$ &
  2 &
  $64\times64$ \\
  
  Head~\cite{ethiopianhead} &
  No &
  400 &
  64 &
  2 &
  No &
  $[0^\circ,90^\circ]$ &
  4 &
  $64\times64$ \\
  
  Toytruck~\cite{Reizenstein21iccv_CO3D} &
  No &
  400 &
  84 &
  2 &
  No &
  $[0^\circ,90^\circ]$ &
  2 &
  $64\times64$ \\
  
  \bottomrule 
  \end{tabular}
  } 
  
  \caption{Summary of the 3D datasets.
  ``Synth.'' indicates if a scene is synthetic (Yes) or real (No).
  RGB-M $=$ RGB-MELON.
  ``Train'' and ``Test'' indicate the number of images in the respective splits.
  $\reporder$ is the replication order used with \methodname.
  For each dataset, we indicate the resolution given at the input of the encoder.}
  \label{tab:datasets}
\end{table}

\begin{figure*}[t]
  \centering
  \includegraphics[width=\linewidth]{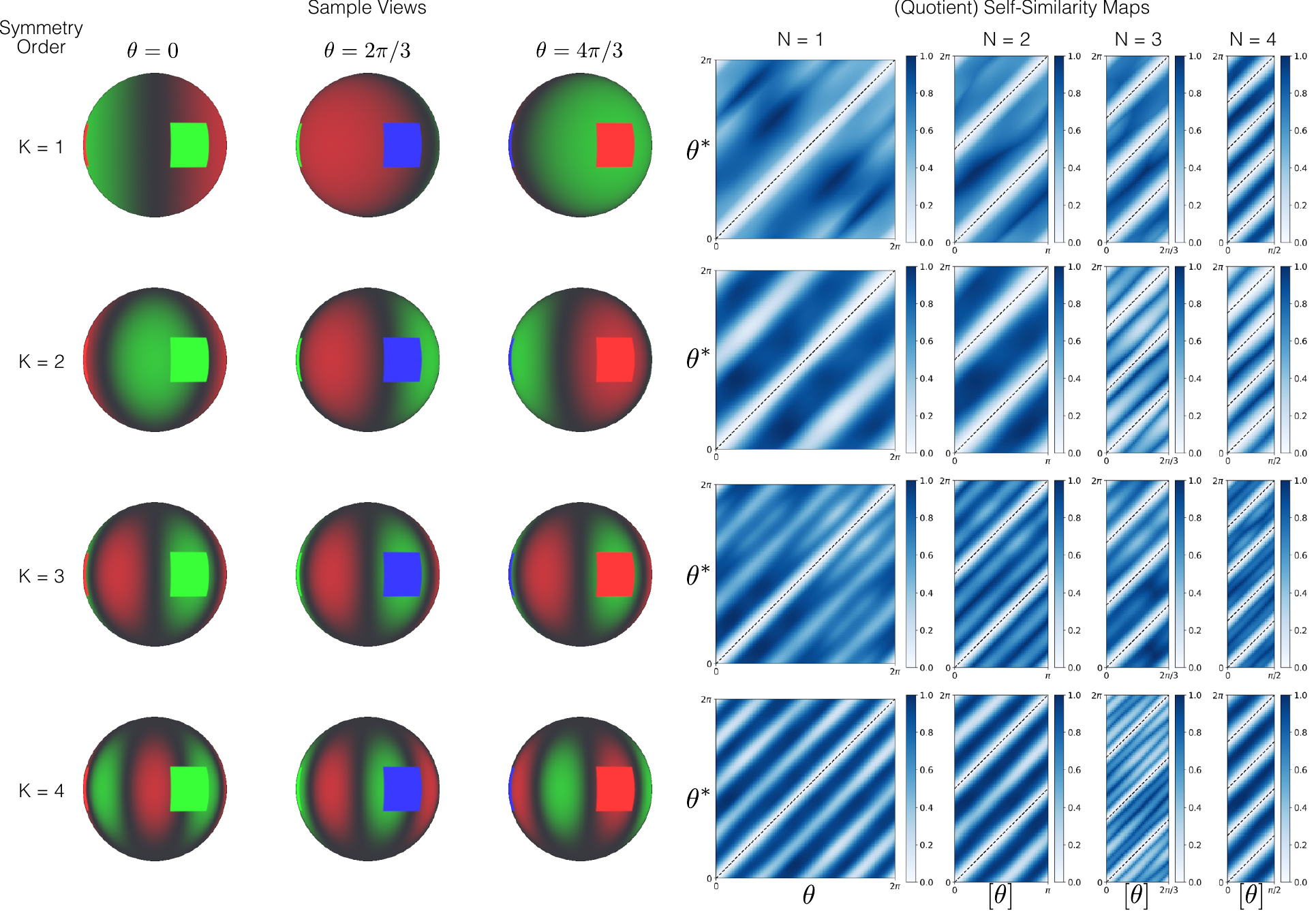}
   \caption{Sample views from the RGB-MELON datasets. The red--green texture is symmetric along the azimuthal direction. In order to make the pose estimation problem well-posed, the symmetry is broken by three red/green/blue squares. All the views are taken from the equator plane. The self-similarity maps and quotient self-similarity maps are plotted on the right for various replication orders $\reporder$.
   }
   \label{fig:sph-dataset}
\end{figure*}

\section{Baselines}

\subsection{COLMAP}

We ran COLMAP with the default configuration for performance evaluations. To ensure fair comparison, we swept parameters, running with different patch window radii (\verb|window_radius|) of 5, 3, 7, 9, 11, and different filtering thresholds of photometric consistency cost (\verb|filter_min_ncc|) of 0.1, 0.01, 0.2, 0.5, and 1.0. We tested all combinations of each.

COLMAP fails when it can not find enough corresponding points between frames. When it finds enough points, it does not always solve all the poses. We report the mean angular error over solved poses in the main text. For fair comparison to MELON and GNeRF, which always predict pose, we also compute a mean over all poses where the error of an unsolved pose is counted as a random guess (\textit{i.e.} $73^\circ$ when the elevation range is $[0^\circ, 90^\circ]$ and $90^\circ$ when it is $[-90^\circ, 90^\circ]$). Results are reported in Table~\ref{tab:colmap-means}.

In general, we note that the high mean angular error of COLMAP is due to large errors on few images. The distribution of angular errors (Fig.~\ref{fig:histo-colmap-errors}) shows a large discrepancy between the median and the mean.

\begin{figure}[!h]
  \centering
  \includegraphics[width=0.6\linewidth]{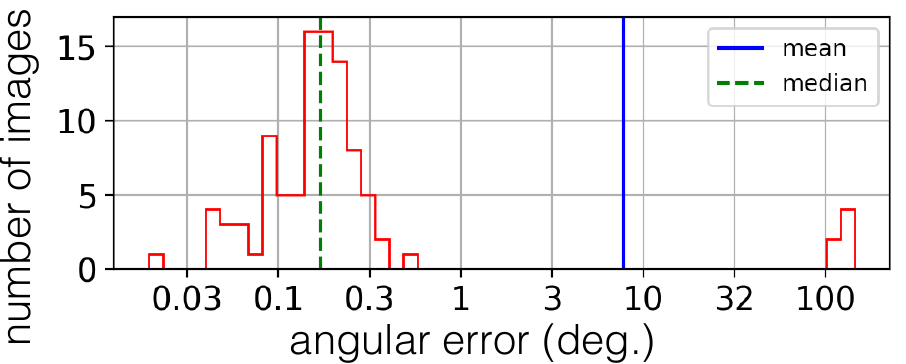}
   \caption{
   Histogram of angular errors obtained with COLMAP on the Lego dataset. 100 images rendered at $128\times128$.
   }
   \label{fig:histo-colmap-errors}
\end{figure}

\begin{table}[h]
  \centering
  \resizebox{\linewidth}{!}{
  \begin{tabular}{l|cccc}
  \toprule
  
   & Mean over& Number of  & Elevation & Mean over \\
  Scene & Solved Poses & Solved Poses & Range &  All Poses \\
  
  \midrule
  Lego & 0.229 & 100/100 & $[0^\circ,90^\circ]$ & 0.229 \\
  Hotdog & 0.328 & 91/100 & $[0^\circ,90^\circ]$ & 6.87 \\
  Chair & 9.71 & 100/100 & $[0^\circ,90^\circ]$ & 9.71 \\
  Drums & 0.467 & 21/100 & $[0^\circ,90^\circ]$ & 57.8 \\
  Mic & 0.860 & 15/100 & $[0^\circ,90^\circ]$ & 62.2 \\
  Ship & 0.292 & 10/100 & $[0^\circ,90^\circ]$ & 65.7 \\
  Materials & 1.05 & 17/100 & $[-90^\circ,90^\circ]$ & 74.9 \\
  Ficus & fails & --- & $[-90^\circ,90^\circ]$ & ---  \\
  
  \bottomrule 
  \end{tabular}
  } 
  
  \caption{Mean angular error of COLMAP poses on the ``NeRF-Synthetic'' scenes.}
  \label{tab:colmap-means}
\end{table}

\subsection{GNeRF}

We run the public version of GNeRF, available at \url{https://github.com/quan-meng/gnerf} with the default parameters. When optimizing the poses in $\text{SO}(3)$, we use $6$-dimensional embeddings to parameterize the camera-to-world rotation matrices $R_i$ in the training and test sets. Based on the value of $R_i$ we set the camera location $t_i$ to
\begin{equation}
    t_i = -R_i\cdot T_i,
\label{eqn:translation-camera-to-world}
\end{equation}
where $T_i$ is the published 3D vector representing the location of the origin in the reference frame of camera $i$. The 3x4 matrix ($R_i$, $t_i$) represents an element of $\text{SE}(3)$.

\subsection{VMRF}
\begin{table*}[t]
  \centering
  \resizebox{\linewidth}{!}{
  \begin{tabular}{lc|ccc|cccc|cccc|cccc}
  \toprule
  & & \multicolumn{3}{c|}{Pose Estimation}
  & \multicolumn{12}{c}{Novel View Synthesis} \\
  
  & & \multicolumn{3}{c|}{Rotation ($^\circ$)~$\downarrow$} 
  & \multicolumn{4}{c|}{PSNR~$\uparrow$} 
  & \multicolumn{4}{c|}{SSIM~$\uparrow$} 
  & \multicolumn{4}{c}{LPIPS~$\downarrow$} \\
  
  \cmidrule(r){3-5} \cmidrule(r){6-9} \cmidrule(r){10-13} \cmidrule(r){14-17}
  
  Scene  & $\reporder$
  & GNeRF & VMRF  & \textbf{\methodname} 
  & GNeRF & VMRF & \textbf{\methodname} & NeRF
  & GNeRF & VMRF & \textbf{\methodname} & NeRF
  & GNeRF & VMRF & \textbf{\methodname} & NeRF \\
  
  \midrule
  
  Lego & 2
  & 3.315 & 1.394  & \textbf{0.059}
  & 22.95 & 25.23 & \textbf{30.22} & 30.44 
  & 0.8493 & 0.8865 & \textbf{0.9525} & 0.9526
  & 0.1630 & 0.1215 & \textbf{0.0627} & 0.0591 \\

  Hotdog & 2
  & --- & ---  & \textbf{3.703}
  & --- & --- & \textbf{27.52} & 35.97 
  & --- & --- & \textbf{0.9428} & 0.9782 
  & --- & --- & \textbf{0.0951} & 0.0384  \\
  
  
  Chair & 4
  & 6.078 & 4.853  & \textbf{2.215}
  & 25.01 & 26.05 & \textbf{30.96} & 32.45
  & 0.8940 & 0.9083 & \textbf{0.9598} & 0.9673
  & 0.1526 & 0.1397 & \textbf{0.0587} & 0.0558 \\

  Drums    & 2
  & 2.769 & 1.284  & \textbf{0.053}
  & 20.63 & 23.07 & \textbf{24.55} & 24.43
  & 0.8628 & 0.8917 & \textbf{0.9121} & 0.9120
  & 0.2019 & 0.1605 & \textbf{0.1096} & 0.1095 \\
  
  Mic & 2
  & 3.021 & 1.394  & \textbf{0.045}
  & 23.68 & 27.63 & \textbf{32.40} & 31.83 
  & 0.9332 & 0.9483 & \textbf{0.9727} & 0.9705
  & 0.1095 & 0.0803 & \textbf{0.0413} & 0.0465 \\
  
  Ship  & 2
  & 31.56 & 16.89  & \textbf{0.210}
  & 17.91 & 21.39 & \textbf{28.27} & 27.71
  & 0.7626 & 0.7998 & \textbf{0.8607} & 0.8553 
  &  0.3628 & 0.2933 & \textbf{0.1704} & 0.1719  \\

  
  
  \bottomrule 
  \end{tabular}
  } 
  
  \caption{Comparison of \methodname to GNeRF and VMRF (results from~\cite{Zhang2022vmrf}). 
  Experiments run with an elevation range of $[-30^\circ,90^\circ]$ as done in VMRF.
  We report the best of five runs.
  $\reporder$ is the replication order used with \methodname.
  }
  \label{tab:ab_initio_vmrf_comparison}
\vspace{-3mm}
\end{table*}


We compare \methodname to VMRF's reported results~\cite{Zhang2022vmrf} by running with the same configuration as in our GNeRF comparison, but with an an expanded elevation range of $[-30^\circ,90^\circ]$, the smallest range reported by VMRF. Results are show in Table \ref{tab:ab_initio_vmrf_comparison}.

\subsection{SAMURAI}

We run the public implementation of SAMURAI, available at \url{https://github.com/google/samurai}. In the fixed initialization scheme, we set all the initial directions to [Center, Above, Center] (North pole). When optimizing the poses in $\text{SO}(3)$, we overwrite the predicted camera locations using~\eqref{eqn:translation-camera-to-world}.

With a manual initialization, each view direction is specified with a triplet [Left/Center/Right, Above/Center/Below, Front/Center/Back]. [Center, Center, Center] is disallowed, giving a total of 26 possible initial directions.

\section{Architecture}

\subsection{Encoder}

Fig.~\ref{fig:encoder} summarizes the architecture of the encoder mapping 2D images to poses. For the 1D experiments, the encoder contains five 1D convolution layers of size $5$ interlaced with SiLU activation functions~\cite{hendrycks2016gaussian}, max-pooling layers and group normalization layers~\cite{wu2018group}.

\begin{figure}[h!]
  \centering
  \includegraphics[width=\linewidth]{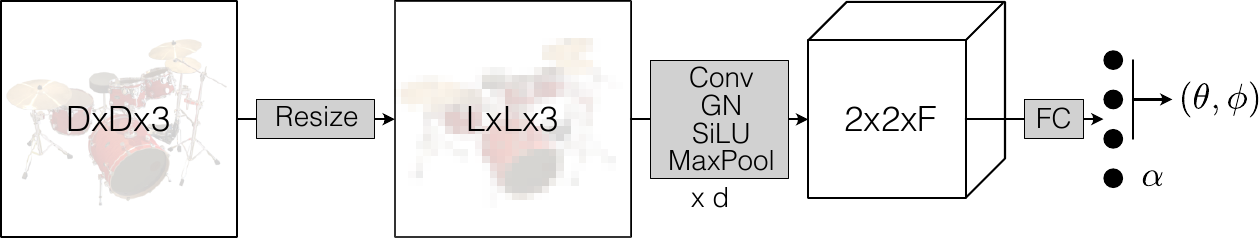}
   \caption{Architecture of the encoder. GN = group normalization~\cite{wu2018group}, SiLU = Sigmoid Linear Unit~\cite{hendrycks2016gaussian}, FC = fully connected layer. We map three of the outputs of the encoder to an azimuth/elevation pair $(\theta,\phi)$ using spherical coordinates. $D$ = input resolution, $L$ = encoder input resolution, $d$ = depth, $F$ = number of features. $\alpha$ is the camera roll, which we only predict for real datasets. Optionally, we can add three output dimensions to predict $T_i$, the location of the origin in the camera reference frame.}
   \label{fig:encoder}
\end{figure}

\subsection{Encoder Ablations}
\label{sec:encoder-ablation}
We perform a series of ablations to explore encoder architecture. We run \methodname on all training views of ``lego'', ``drum'', ``ficus'' and ``ship'' at $400\times400$ resolution and $\reporder=2$. We observe in our primary experiments that when poses converge, they tend to do so very quickly, often in the first thousand steps. We therefore train each configuration for 10k steps, but repeat 10 times to find min, max and mean metrics. We compute metrics as in other experiments over all test views. We choose a primary configuration of $L=64$, $F=32$, $d=4$ varying one parameter at a time in our experiments.

\begin{figure}
    \centering
    \includegraphics[width=\linewidth]{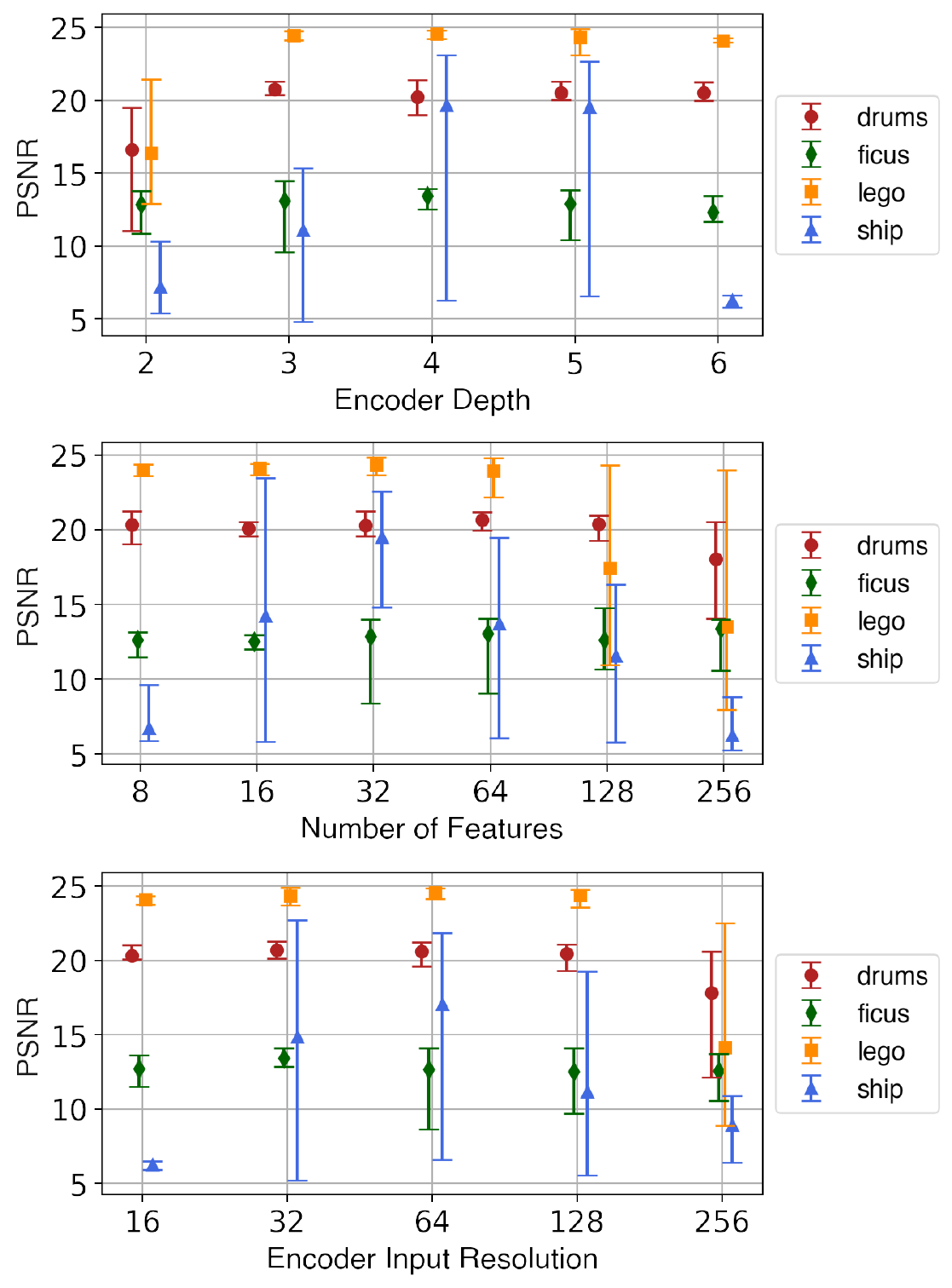}
    \caption{PSNR of novel view synthesis vs. encoder depth $d$, number of features $F$, and input resolution $L$, for four ``NeRF-Synthetic'' scenes. We report min, max and mean values over 10 runs.}
    \label{fig:encoder-ablation}
\end{figure}

Fig.~\ref{fig:encoder-ablation} shows the result of sweeping feature count $F$ from 8 to 256 and encoder depth $d$ from 2 to 6 layers. We observe optimal values at 4 layers, and a feature count of 32. 

Fig.~\ref{fig:encoder-ablation} also shows performance sweeping encoder input resolutions $L$ from $16\times16$ to $256\times256$. We observe that certain datasets appear to work better at different resolutions, with optimal values typically ranging from $32\times32$ to $128\times128$. We therefore choose per-scene values in our primary experiments.

\subsection{Replication Order}
\begin{figure}[ht]
    \centering
    \includegraphics[width=\linewidth]{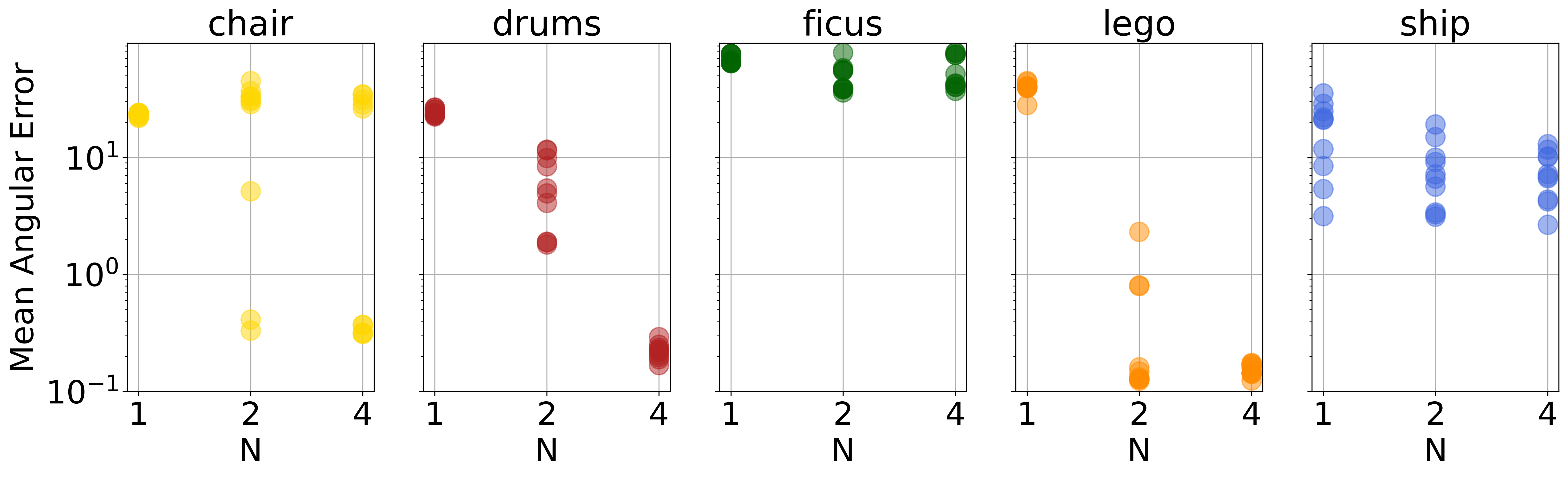}
    \caption{Scatter plot of the mean angular error vs. replication order $\reporder$ for various ``NeRF-Synthetic'' scenes. We run each scene 10 times to 10k steps.}
    \label{fig:replication-order}
\end{figure}
To determine the effect of varying the replication order, we run with the same base configuration as Sec.~\ref{sec:encoder-ablation}, but varying replication order over $\reporder = \{1, 2, 4\}$. We run \methodname 10 times to 10k steps on five scenes and plot the results in Fig.~\ref{fig:replication-order}. We observe a clear bi-modal behavior of convergence / non-convergence, and likelier convergence with higher $\reporder$, consistent with our analysis of Sec.~3.4. The scenes ``ficus'' and ``ship'' typically take longer than 10k steps to converge, and thus are not well illustrated by this study.

\subsection{Neural Radiance Field}
We use the standard NeRF architecture of \cite{Mildenhall20eccv_nerf}, an MLP with 8 layers of 256 features, and one skip connection to layer 4. Our density and RGB branches use a single layer of 128 channels. We use a single MLP, sampled at 96 stratified points, plus 32 additional importance sampled points, guided by the initial stratified samples. We keep the $t_f - t_n$ span fixed at 3.0 for baseline NeRF runs with ground truth cameras. For \methodname runs, we linearly interpolate from 1.0 to 3.0 during the first 10k steps, which we found to aid convergence. We ignore view dependence for the first 50k steps by providing random view directions to the network. After 50k steps, we anneal a maximum of 4 frequency bands of directional encoding, until training completion at 100k steps. We use the Adam optimizer \cite{kingma2014adam} with a learning rate of $10^{-4}$, $\beta_1=0.9, \beta_2=0.999$ and $128$ pixels per batch.

\section{Additional Results}

\subsection{Lego Scene}

We show the self-similarity map and the regions of attraction of the Lego scene for a fixed elevation and a variable reference pose $z^*$ in Fig.~\ref{fig:ssm-roa-lego-fixed-elevation}.

\begin{figure}[ht]
  \centering
  \includegraphics[width=\linewidth]{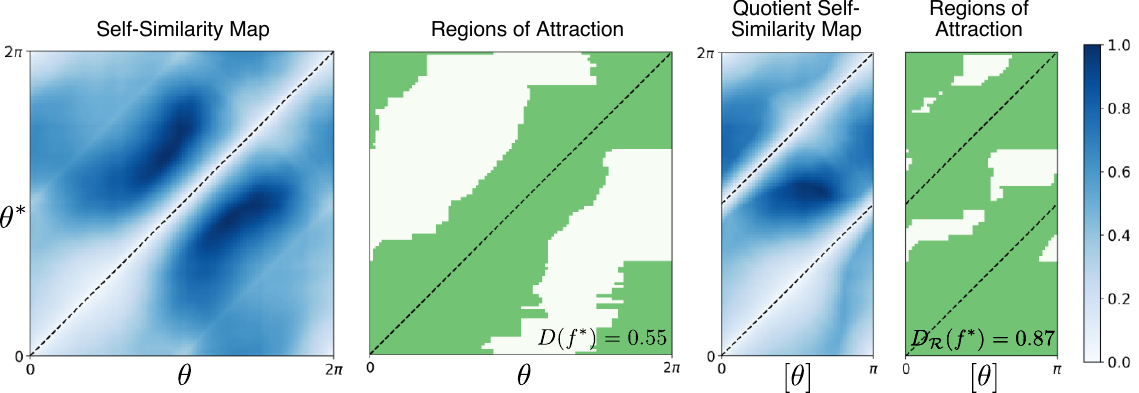}
   \caption{
   We generate 100 views of the ``lego'' scene with from an elevation $\phi\simeq 7^\circ$ and an azimuth $\theta\in[0^\circ,360^\circ]$. The (normalized) self-similarity map $\degen_{f^*}$ shows that the views taken from opposite sides look similar, due to the presence of rotationally symmetric elements in the bulldozer. The regions of attraction $\conv_{f^*}(\theta^*)$ are colored in green. They only partially cover the latent space ($D(f^*)=0.55$, higher is better, see~\eqref{eq:difficulty}). With the equivalence relation $\relation=\relation_2$, the regions of attraction cover the quotient space $\latentspace/\relation$ almost entirely ($D_\relation(f^*)=0.87$, see~\eqref{eq:difficulty-rep}). Black dotted lines correspond to $\theta^*\in[\theta]$.
   }
   \label{fig:ssm-roa-lego-fixed-elevation}
\vspace{-3mm}
\end{figure}

\subsection{Comparison to VMRF}

In our comparison to VMRF, the predicted elevation is constrained to the range $[-30^\circ, 90^\circ]$, as done in~\cite{Zhang2022vmrf}. \methodname achieves higher reconstruction metrics on all the ``NeRF-Synthetic'' scenes (Table~\ref{tab:ab_initio_vmrf_comparison}).

\subsection{Number of Views}

We show qualitative reconstructions obtained on the ``lego'' dataset with a varying number of views in the training set in Fig.~\ref{fig:num-views-qual}.

\subsection{NeRF in the Wild}

We report quantitative metrics obtains on the real datasets in Table~\ref{tab:nerf_in_the_wild_qualitative} and qualitative reconstructions in Fig.~\ref{fig:nerf_in_the_wild_so3_se3}.

\begin{table}[ht]
  \centering
  \resizebox{\linewidth}{!}{
  \begin{tabular}{l|cc|cc|cc|cc|c}
  \toprule
  
  &
  \multicolumn{2}{c|}{GNeRF} &
  \multicolumn{2}{c|}{SAMURAI*} &
  \multicolumn{2}{c|}{\textbf{\methodname}} &
  \multicolumn{2}{c|}{SAMURAI} &
  NeRF \\
  
  Scene &
  Rot. ($^\circ$) & PSNR &
  Rot. ($^\circ$) & PSNR &
  Rot. ($^\circ$) & PSNR &
  Rot. ($^\circ$) & PSNR &
  PSNR \\
  
  \midrule
  
  Cape &
  56 & 16.3 &
  56 & 12.9 &
  \textbf{1.5} & \underline{19.5} &
  5.1 & \textbf{20.6} &
  19.1\\
  
  Head &
  42 & 14.5 &
  46 & 19.0 &
  \textbf{2.0} & 20.6 &
  7.1 & \underline{23.5} &
  \textbf{24.3} \\
  
  Toytruck &
  45 & 13.6 &
  54 & 15.6 &
  \textbf{2.5} & \underline{23.5} &
  3.4 & 22.2 &
  \textbf{26.2} \\
  
  \bottomrule 
  \end{tabular}
  } 
  
  \caption{Mean angular error on predicted poses (training set) and quality of novel view synthesis (test set). SAMURAI* uses a fixed initialization of the poses at the North pole while SAMURAI uses a manual rough initialization. ``NeRF'' is our implementation using published poses, provided by COLMAP. SAMURAI can achieve a better reconstruction thanks to a more flexible 3D representation (shape, BRDF and per-camera illumination), in spite of less accurate predicted poses. As ground truth poses are provided by COLMAP, angular errors may be noisy estimates of the true pose accuracy.}
  \label{tab:nerf_in_the_wild_qualitative}
\end{table}

\subsection{Noise Sensitivity}

We compare \methodname to COLMAP and GNeRF on noisy datasets. We add pixel-independent Gaussian noise of variance $\sigma^2$ to the training images and evaluate the mean angular error and novel view synthesis accuracy of competing methods in Table~\ref{tab:ab_initio_vmrf_comparison}. We show a qualitative comparison between \methodname and GNeRF in Fig.~\ref{fig:noise-sweep-gnerf}.

\begin{table}[h]
  \centering
  \resizebox{\linewidth}{!}{
  \begin{tabular}{l|ccc|cc}
  \toprule
  \multicolumn{1}{c|}{Noise} & \multicolumn{3}{c|}{Pose Estimation}        & \multicolumn{2}{c}{Novel View Synthesis} \\
  \multicolumn{1}{c|}{Std. Dev.} & \multicolumn{3}{c|}{Rotation ($^\circ$)$\downarrow$} & \multicolumn{2}{c}{PSNR$\uparrow$} \\
  \cmidrule(r){2-4} \cmidrule(r){5-6}
  $\sigma$ & COLMAP & GNeRF &   \textbf{\methodname} & GNeRF & \textbf{\methodname}\\
  \midrule

  

  

  0.125    & 
   fails & 2.351 & \textbf{0.119} &  
  30.15 & \textbf{30.25} \\

  0.25    & 
   fails & 4.271 & \textbf{0.190} &
  27.32 & \textbf{28.62} \\  

  0.5    & 
   fails & 5.883 & \textbf{0.363} &  
  24.95 & \textbf{26.66} \\
  
  1.0    & 
   fails & 7.557 & \textbf{1.784} &  
  22.90 & \textbf{24.20} \\
  
  2.0    & 
   fails & 31.06 & \textbf{8.416} &  
  17.12 & \textbf{19.99} \\
  
  \bottomrule 
  \end{tabular}
  } 
  
    \caption{Pose estimation accuracy and novel view synthesis quality of competing methods on the noisy ``lego'' scene.
    We report the best of three runs.
  }
  \label{tab:image_noise_sweep}
\end{table}

\begin{figure}[h]
  \centering
  \includegraphics[width=\linewidth]{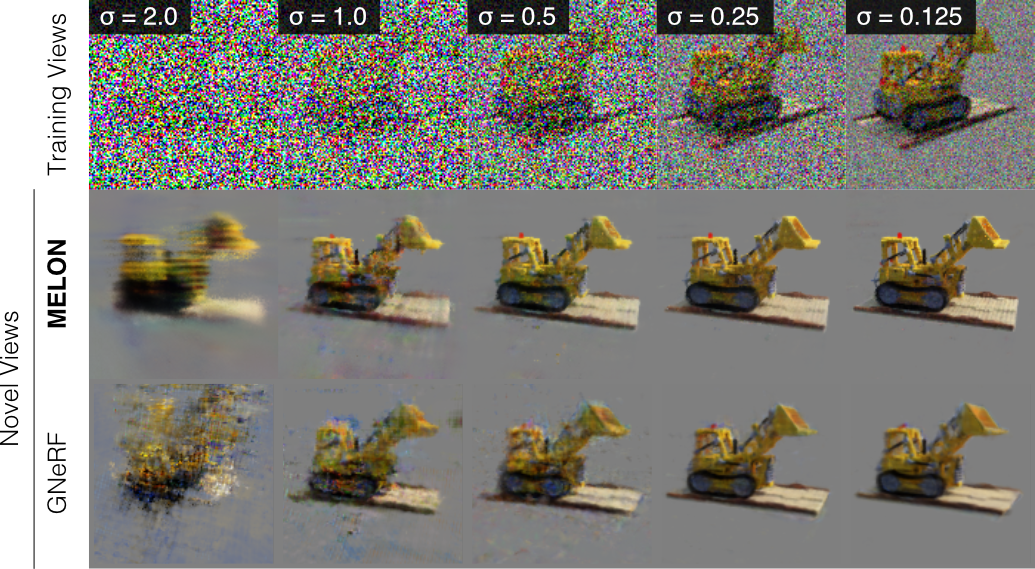}
   \caption{Novel view synthesis from noisy unposed images ($128\times 128$). Comparison between \methodname and GNeRF.}
   \label{fig:noise-sweep-gnerf}
\end{figure}

\begin{figure*}
  \centering
  \includegraphics[width=\linewidth]{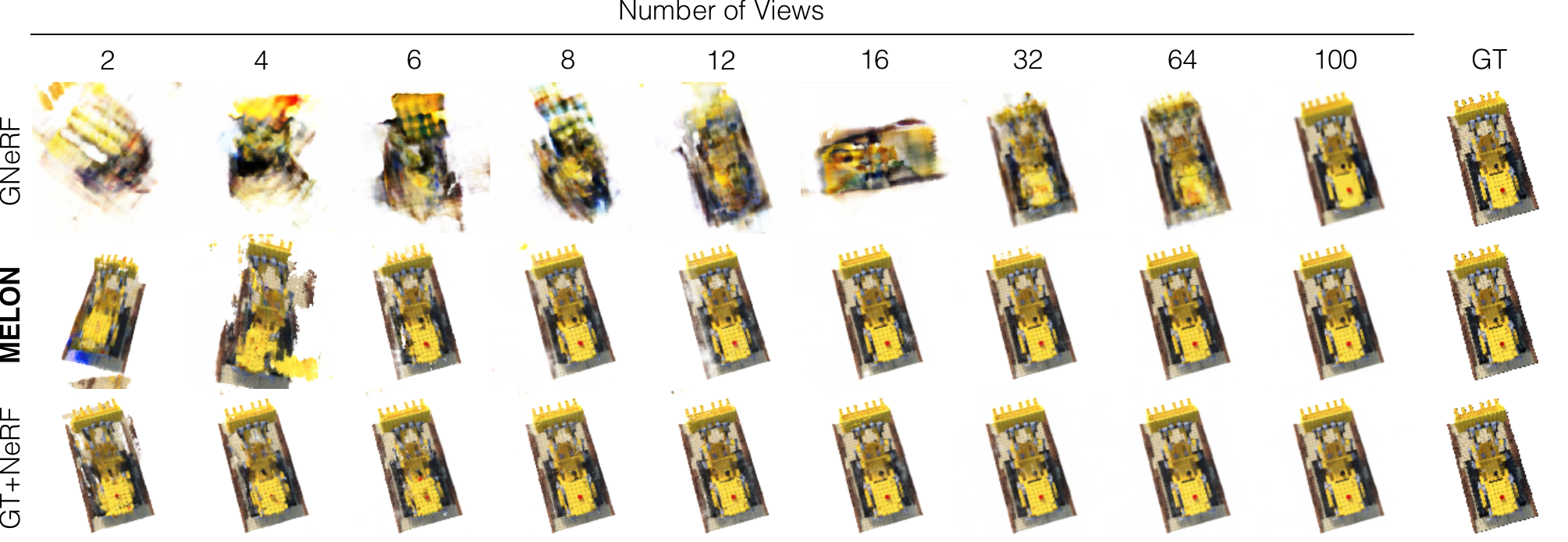}
   \caption{Qualitative reconstructions for various numbers of training views on the ``lego'' scene.}
   \label{fig:num-views-qual}
\end{figure*}

\begin{figure*}
  \centering
  \includegraphics[width=0.8\linewidth]{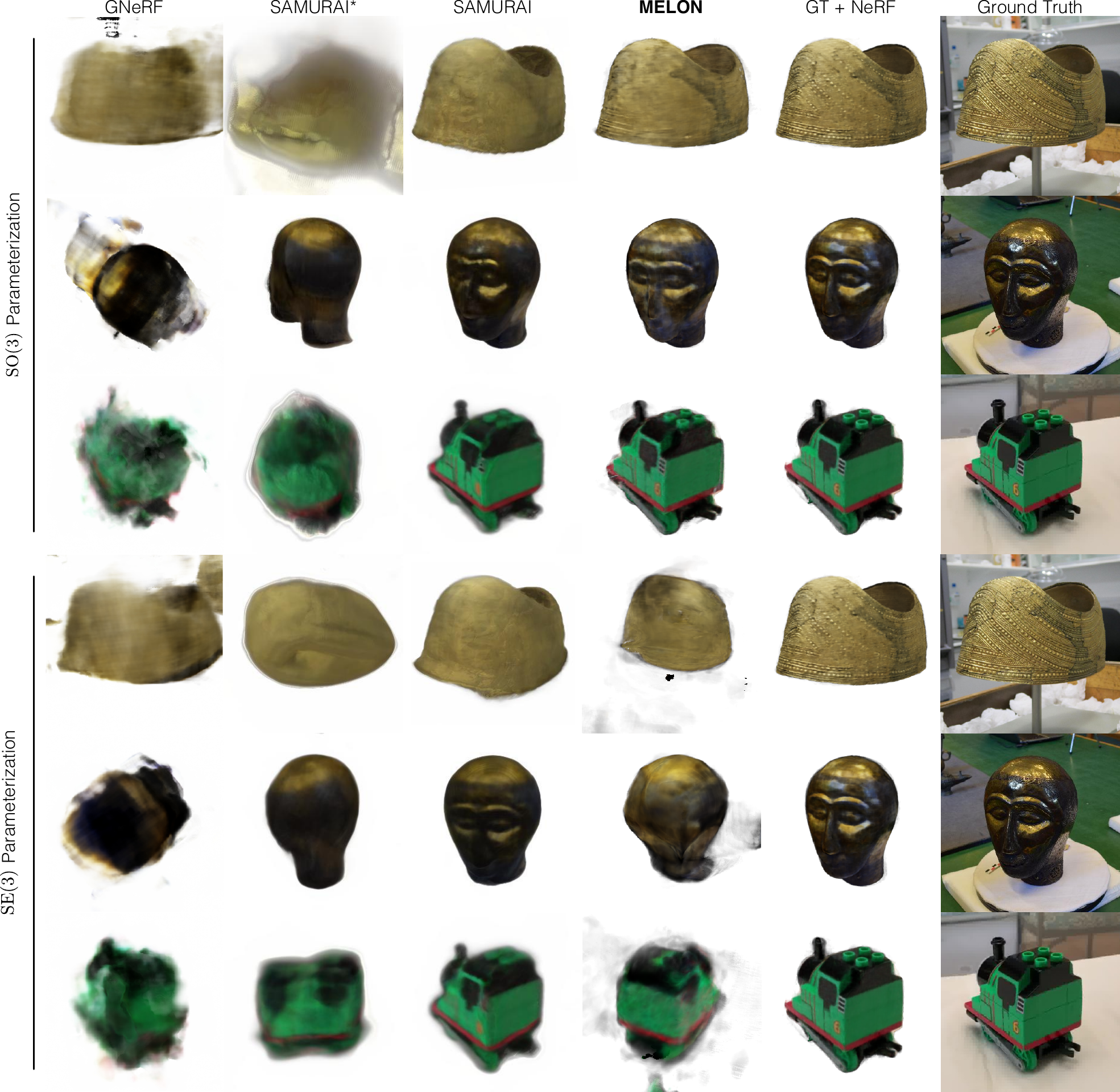}
   \caption{Additional qualitative results on the real datasets. Ground truth (GT) poses are provided by COLMAP. SAMURAI* uses a fixed initialization of the poses at the North pole while SAMURAI uses a manual coarse initialization. With the $\sothree$ parameterization, the object-to-camera distances and the camera in-plane translations are assumed to be known.}
   \label{fig:nerf_in_the_wild_so3_se3}
\end{figure*}

{\small
\bibliographystyle{ieeenat_fullname}
\bibliography{nerf,non-nerf}
}
}{
\maketitle

\begin{abstract}
Neural radiance fields enable novel-view synthesis and scene reconstruction with photorealistic quality from a few images, but require known and accurate camera poses. Conventional pose estimation algorithms fail on smooth or self-similar scenes, while methods performing inverse rendering from unposed views require a rough initialization of the camera orientations. The main difficulty of pose estimation lies in real-life objects being almost invariant under certain transformations, making the photometric distance between rendered views non-convex with respect to the camera parameters. Using an equivalence relation that matches the distribution of local minima in camera space, we reduce this space to its quotient set, in which pose estimation becomes a more convex problem. Using a neural-network to regularize pose estimation, we demonstrate that our method -- MELON -- can reconstruct a neural radiance field from unposed images with state-of-the-art accuracy while requiring ten times fewer views than adversarial approaches.
\end{abstract}

\section{Introduction}
\label{sec:intro}

Neural rendering methods have demonstrated wide ranging application from novel-view synthesis~\cite{Mildenhall20eccv_nerf}, to avatar generation for virtual reality ~\cite{Gafni21cvpr_DNRF}, to 3D reconstruction of molecules from microscope images ~\cite{Levy2022cryoai}.
Neural radiance fields (NeRFs)~\cite{Mildenhall20eccv_nerf} represent radiance with a neural field that reproduces the geometric structure and appearance of a scene, allowing the use of gradient descent to reconstruct a set of input images.
NeRFs have gained wide acceptance for their robust ability to generate novel-views of a 3D scene, but must be trained with known camera poses, limiting their use to cases where camera pose is known or can be inferred from other methods.
Our method seeks to estimate camera pose and radiance field simultaneously, eliminating the requirement of known camera poses.

Classical structure-from-motion algorithms (SfM), such as COLMAP~\cite{Schonberger2016sfmrevisited}, compute the relative camera pose between  views by detecting and matching salient points in the scene and minimizing re-projection error.
Despite being widely used, SfM commonly fails on scenes containing many view-dependent effects or few textures and edges.
Other methods \cite{Lin21iccv_BARF, Wang21arxiv_NeRFminusminus, Chng2022garf} jointly estimate a neural radiance field and camera parameters  using gradient descent, but require an approximate initialization of camera parameters, and none work in a $360^\circ$ object centered configuration with random initialization. Adversarial approaches \cite{Henzler2019platonicgan, NguyenPhuoc2019hologan, Niemeyer2021giraffe, Niemeyer21cvpr_GIRAFFE, Chan21cvpr_piGAN, Meng21iccv_GNeRF} jointly train a 3D generator representing the scene, and a discriminator that attempts to differentiate input images from generated images.
These methods, however, suffer from large input data requirements, hyper parameter sensitivity, complex training schedules, and high computation times.

\begin{figure}[t]
  \centering
  \includegraphics[width=\linewidth]{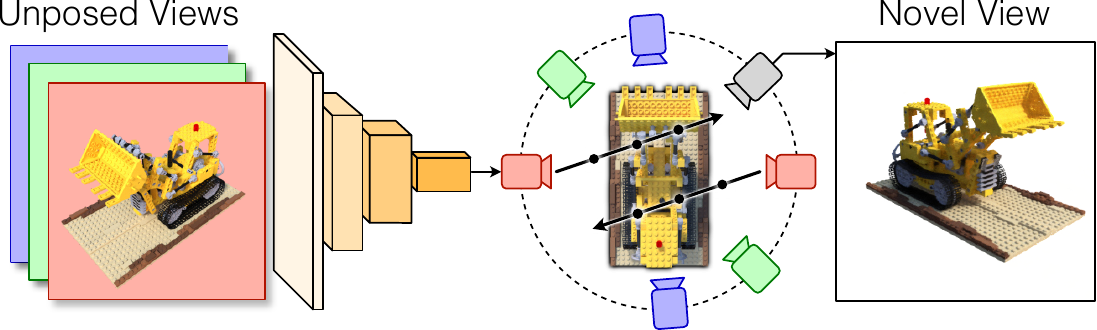}
   \caption{\methodname infers a neural radiance field from an object-centered dataset.
   A dataset-specific encoder maps each image to a predicted pose $R_i$. Given an equivalence relation in camera space, we render rays from all the poses in the equivalence class $[R_i]$. The modulo loss only penalizes the smallest L2 distance from the ground truth color. At evaluation time, the neural field can be used to generate novel views.
   }
   \label{fig:teaser}
\vspace{-3mm}
\end{figure}
We introduce \methodname (Modulo-Equivalent Latent Optimization of NeRF), an encoder-decoder-based method that infers a neural representation of a scene from unposed views (Fig.~\ref{fig:teaser}).
Our method simultaneously trains a CNN encoder that maps images to camera poses in $\sothree$, and a neural radiance field of the scene.
\methodname does not require CNN pre-training, and is able to infer camera pose in object centered configurations entirely ``\emph{ab-initio}'', \textit{i.e.} without any pose initialization whatsoever.
To cope with the presence of local minima in the low-dimensional latent space $\sothree$, we introduce a novel \textit{\lossname} that replicates the encoder output following a pre-defined structure.
We formalize this replication with an equivalence relation defined in the space of camera parameters, and show this enables the encoder to operate in a quotient set of the camera space.
We introduce a one-dimensional toy problem that mimics the challenges posed by 3D inverse rendering to better illustrate our method.

We identify the following contributions:
\begin{itemize}
\item We introduce a gradient-descent-based algorithm to train a neural radiance field from entirely unposed images, demonstrating competitive reconstruction metrics on a variety of synthetic and real datasets.
\item We show that \methodname is robust to noise and can perform pose estimation and novel-view synthesis from as few as six unposed images.
\item We introduce 1D and 3D datasets of almost symmetric objects which can be used as challenging examples for pose estimation and inverse rendering.
\end{itemize}

Our code will be released upon publication.

\section{Related Work}
\hyphenation{iNeRF}



\label{sec:relatedwork}
\paragraph{Neural Radiance Fields~(NeRF)}~\cite{Mildenhall20eccv_nerf} use classical volume rendering~\cite{Kajiya1984rtv} to compute the RGB values $\nerfcolor$ for each pixel~$\pixel$ from samples taken at points $\position$ along a ray of direction $\viewdir$. The ray direction is determined by the camera parameters $R$ and the pixel location.
These samples are computed using learned radiance ($\radiance$) and density ($\density$) fields.
The volume rendering equation takes the form of an integral along a ray between near and far view boundaries $(t_n,t_f)$:
\begin{equation}
\nerfcolor(\pixel~;\radiance,\density,R) = \int_{t_n}^{t_f} T(t)\density(\position(t))\radiance(\position(t),\viewdir)\diff t
\label{eqn:nerf}
\end{equation}
where the transmittance is given by
\begin{equation}
\label{eqn:sample_weight}
T(t)= \exp\left(-\int_{t_n}^t\density(\position(s))\diff s\right).
\end{equation}

Neural Radiance Fields have proven capable of high-quality novel view synthesis and have been a recent focus in the research community.
They have been extended in many ways including training from noisy HDR images \cite{Mildenhall2022rawnerf}, grid-based representations for faster training and inference~\cite{Sun2022dvgo, Muller2022ingp, Wang2022plenoctrees, Reiser21iccv_KiloNeRF}, training with limited or single views \cite{Yu21cvpr_pixelNeRF, Jain21iccv_DietNeRF, Xu2022sinnerf}, decomposition into sub-fields~\cite{Rebain20arxiv_derf, Reiser21iccv_KiloNeRF}, unconstrained images in the wild \cite{MartinBrualla21cvpr_nerfw}, deformable fields \cite{Park21iccv_nerfies, Noguchi21iccv_NARF, Raj2021pva}, and video~\cite{Li2022dynerf}. 
Most of these techniques however assume camera poses to be given, and automatically inferring them during training remains an open research question.

\textbf{Structure-from-Motion.}
Reconstruction from unposed images has been a major thrust of research in the structure from motion community for many years~\cite{Ozyesil2017sfmsurvey, Iglhhaut2019sfmreview}. 
COLMAP~\cite{Schonberger2016sfmrevisited} is currently the most commonly used method to infer pose when unavailable.
Though robustly usable for a number of cases, COLMAP relies on SIFT feature matching and can fail in cases where SIFT features cannot be found, such as scenes with limited edges, corners and texture, or view-dependent effects like reflections, refractions or specular surfaces.

\textbf{Neural Representations without Poses.}
Pose estimation of neural scenes can be divided into those which assume a known scene, and those which simultaneously reconstruct a scene along with poses. 
Of those assuming a known scene, iNeRF~\cite{YenChen20iros_iNeRF}, further extended by~\cite{Lin2022pin}, demonstrated that pose could be recovered with stochastic gradient descent and ``interest region sampling'', when initialized within $\pm40^\circ$ of ground truth orientation.

NeRF$\shortminus\shortminus$~\cite{Wang21arxiv_NeRFminusminus} was the first to co-optimize a neural field and 6-DOF camera poses simultaneously.
To avoid local minima, the neural field was re-initialized mid-training, and results were only shown for forward-facing scenes.
BARF~\cite{Lin21iccv_BARF} proposed a coarse-to-fine positional encoding annealing strategy~\cite{Park21iccv_nerfies} and GARF~\cite{Chng2022garf} argued that Gaussian activation functions were more robust for pose estimation. SCNeRF~\cite{Jeong2021scnerf} extended the parameterization to include camera instrinsics and, like SPARF~\cite{truong2022sparf}, incorporated multi-view geometric constraints. These methods avoid the need to re-initialize the field and demonstrated results for non forward-facing scenes, but require a rough initialization of the camera poses. SAMURAI~\cite{Boss2022samurai} simultaneously optimizes several cameras with a novel camera parameterization and tackles non forward-facing scenes but requires poses to be initialized to one of 26 possible directions. NoPe-NeRF~\cite{Bian2022nopenerf} introduces additional supervision from an off-the-shelf monocular depth estimation network and novel losses that encourage 3D consistency between the scene and successive frames in a video sequence.
All these techniques require initialization within some fixed bound of ground truth pose, or a series of views known to be close in pose to one another.

Approaches like iMAP~\cite{Sucar21iccv_iMAP} or NICE-SLAM~\cite{Zhu2022niceslam} perform reconstruction from hand-held camera, but requires depth information and ignore view-dependence in the neural field.

\textbf{Adversarial Approaches }\cite{Henzler2019platonicgan, NguyenPhuoc2019hologan, Niemeyer2021giraffe, Chan21cvpr_piGAN} jointly supervise a generator containing a differentiable neural field and a discriminator.
Such techniques enable the reconstruction of a radiance field while bypassing the pose estimation problem.
GNeRF~\cite{Meng21iccv_GNeRF} trains on a single scene by supplying image crops to the discriminator, increasing the effective data set size. It co-trains a pose encoder supervised with renderings of the neural scene, which is then used to estimate ground truth image poses.
Using a complex training schedule GNeRF, is able to reconstruct a single scene from completely unposed images.
VMRF~\cite{Zhang2022vmrf} refined this work by using a pre-trained feature detector, differentiable unbalanced optimal transport solver and relative pose predictor to provide extra supervision achieving better reconstruction and pose estimation than GNeRF.

Although such techniques enable the reconstruction of a photo-realistic 3D field from unposed images, they require large datasets for training the discriminator, complex training schedules, and suffer the inherent instabilities and hyper-parameter sensitivities of GANs.

\textbf{Supervised Pose Regression} methods approach the pose estimation (or ``camera calibration'') problem as a regression problem, where a mapping from image space to pose space needs to be approximated \cite{Hu2020sss, lian2022end, kendall2015posenet, kendall2017geometric, mahendran20173d, su2015render, li2019monocular, Lopez2019dsi} and \cite{sattler2019understanding} analyzed the limits of CNN-based methods.
Focusing on the degenerate case of symmetric objects, Implicit-PDF~\cite{murphy2021implicit} showed that an image can be mapped to a multimodal posterior distributions stored in a coordinate-based representation and RelPose~\cite{zhang2022relpose} introduced an energy-based approach to estimate joint distribution over relative viewpoints in $\sothree$. SparsePose~\cite{sinha2022sparsepose} trains a coarse pose regression model on a large dataset and refines camera poses in $\sethree$ in an auto-regressive manner.
However, these approaches require training over a large data set with known poses and do not provide a neural field of the scene.

Previous works proposed ways to build rotation-equivariant networks \cite{Deng2021vectorneurons, elesedy2021provably, batzner20223, fuchs2020se, chen2021equivariant, esteves2018learning, esteves2019equivariant, nasiri2022unsupervised}.
Such symmetry-aware networks showed improved generalization properties for processing point clouds or spherical images but do not ensure $\text{SO}(3)$-equivariance for pose estimation since no group action for the group $\text{SO}(3)$ can be defined on the set of 2D images~\cite{Klee2022iip}.

\textbf{Unsupervised Approaches }use an encoder--decoder architecture mapping the image space to an $\text{SO}(3)$-valued latent space and mapping this space back to the image space with a group-equivariant generative model \cite{Falorsi2018ehv}.
Spatial-VAE \cite{bepler2019explicitly} demonstrated translation and in-plane rotation estimation for 2D images using a neural-based 2D generative model in the decoder, while TARGET-VAE \cite{nasiri2022unsupervised} proposed to use a translation and in-plane rotation equivariant encoder.
SaNeRF \cite{Chen2022san} runs classical COLMAP SIFT feature detection and mapping, but regresses relative poses from a triplet of images fed to a CNN.
As it uses SIFT feature matching, this method is not suitable for objects containing large texture-less regions or repeated structures.

Approaches from the Cryo-Electron Microscopy (Cryo-EM) community \cite{Levy2022aih, Levy2022cryoai} recently showed the possibility of using an auto-encoding approach to reconstruct a 3D neural field in Fourier domain~\cite{zhong2021cryodrgn, shekarforoush2022residual} from unposed 2D projections.
These methods do not require any initialization for the poses and have been shown to work on datasets with high levels of noise.
The image formation model in cryo-EM is orthographic and does not involve opacity, allowing faster computation with the Fourier Slice Theorem.
Building on these approaches, our method copes with a more complex rendering equation that incorporates perspective and opacity.

\begin{figure}[t]
  \centering
  \includegraphics[width=\linewidth]{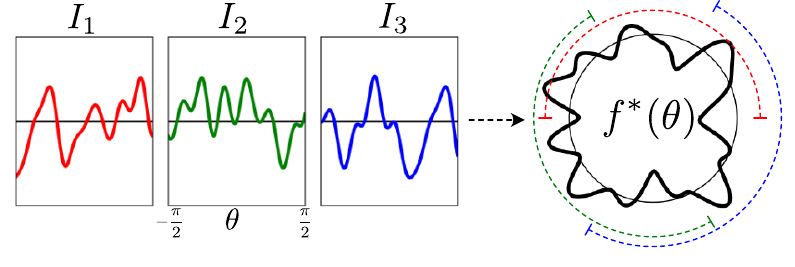}
   \caption{1D inverse rendering. We are given a set of crops $I_i$ from an unknown function $f^*(\theta)$ mapping $\sotwo$ to $\mathbb{R}$. Each crop has size $\pi$ and is centered around an unknown angle $\theta_i^*$. The goal is to recover the function (and the angles) given the crops.
   }
   \label{fig:method-1d-labeled}
\vspace{-3mm}
\end{figure}

\section{Methods}

We are given a set of input images $\{\image\}_{i=1 \ldots n}$ which we assume to be renderings of an unknown density $\density^*$ and view-dependent radiance $\radiance^*$ field from unknown cameras $\{\pose^*\}_{i=1,\ldots,n}$, following~\eqref{eqn:nerf}:
\begin{equation}
    \forall i,\forall\pixel\in\plane,\quad\image(\pixel)=\nerfcolor(\pixel~; \shared^*,\pose^*)
\label{eqn:three-d-forward-model}
\end{equation}
where $\plane$ is a discrete set of pixel coordinates and $\shared^*=(\radiance^*,\density^*)$.
We assume that all cameras have known intrinsics and point towards the origin of the scene from a known distance.
Each camera orientation $R_i\in\sothree$ can therefore be described with three numbers: azimuth $\azim\in\sotwo$, elevation $\elev\in[-\pi/2,\pi/2]$ and roll $\roll\in\sotwo$. We use the symbol $\sim$ to represent an element of $\sothree$ in this coordinate system.
We define (external) additive and multiplicative operations such that, for $\theta_0\in\sotwo$, $\lambda\in\mathbb{R}$ and $R\sim(\azim,\elev,\roll)$,
\begin{equation}
    \theta_0+R\sim(\theta_0+\azim,\elev,\roll)\quad;\quad\lambda R\sim(\lambda \azim,\elev,\roll).
\end{equation}

We use a neural radiance field parameterized by $\weightsshared$ to represent the predicted function $\shared_\weightsshared=(\radiance_\weightsshared,\density_\weightsshared)$. The goal of \textit{3D inverse rendering with unknown poses} is to approximate $f^*$ with $\shared_\weightsshared$ and $R_i^*$ with $R_i$, up to a global rotation.
While $\weightsshared$ is shared, each image is associated with a unique $\pose$, called a \textit{latent variable} living in the \textit{latent space} $\sothree$. The low dimensionality of this space induces the presence of local minima~\cite{bray2007statistics, dauphin2014identifying} and makes pose estimation difficult to solve by gradient descent.

\subsection{One-dimensional Toy Problem}

We propose a simpler 1D version of the above problem to aid us in understanding the challenges existing in 3D space (Fig.~\ref{fig:method-1d-labeled}). 
Assume we are given $n$ crops $\{\image\}_{i=1,\ldots,n}$ of width $\pi$ of an unknown one-dimensional function $\torus^*$ mapping $\sotwo$ to $\mathbb{R}$. Each crop $\image$ is centered around an unknown angle $\ang^*$:
\begin{equation}
    \forall i,\forall\pix\in\segment,\quad\image(\pix)=\nerfcolor(\pix~;\torus^*,\ang^*)=\torus^*(\ang^*+\pix)
\label{eqn:one-d-forward-model}
\end{equation}
where $\segment=[-\pi/2,\pi/2]\subset\sotwo$. Our objective is to recover the function $f^*$ given the crops $\image$, without knowing the set of angles $\{\theta_i^*\}$.

The objective of \textit{1D inverse rendering with unknown angles} is to approximate $\torus^*$ with a coordinate-based neural network $\torus_\weightsshared$ and $\theta_i^*$ with $\theta_i$, up to a global rotation.
In this case, the latent variables are the angles $\ang$ belonging to the latent space $\sotwo$.

\subsection{Amortized Inference of Latent Variables}

In both problems, a set of latent variables $\latent$ must be estimated ($\pose$ in 3D, $\ang$ in 1D) jointly with a parameter $\weightsshared$. Instead of minimizing
$\norm\big{\nerfcolor(.~;\torus_\weightsshared,z_i)-\image(.)}_{2}^2$ over each $\latent$ independently, we aim to find an \textit{encoder} $\enc$ mapping observations $\image$ (images or crops) to their associated latent variables $\latent^*$. The encoder is parameterized by a neural network with weights $\weightsenc$ (see details on the architecture in the supplements). The objective function of our optimization is therefore
\begin{equation}
    \loss(\weightsshared,\xi)=\sum_{i=1}^n\norm\big{ \nerfcolor(.~;\shared_\weightsshared,\enc_\xi(\image)) - \image(.)}_2^2,
\label{eqn:generic-loss}
\end{equation}
where $\Vert.\Vert_2$ is the L2 norm on the plane $\plane$ or segment $\segment$. We say that the inference of the latent variables is \textit{amortized} over the size of the dataset, meaning that the estimation of the latent variables for the dataset is seen as a unique inference problem instead of $n$ independent problems. In particular, if $\enc$ is accurate on a given observation $\image$, it will likely be accurate on similar observations provided that it is likely to be \textit{smooth} in observation space.

\subsection{Self-Similarity Map}

\begin{figure*}[t]
  \centering
  \includegraphics[width=\linewidth]{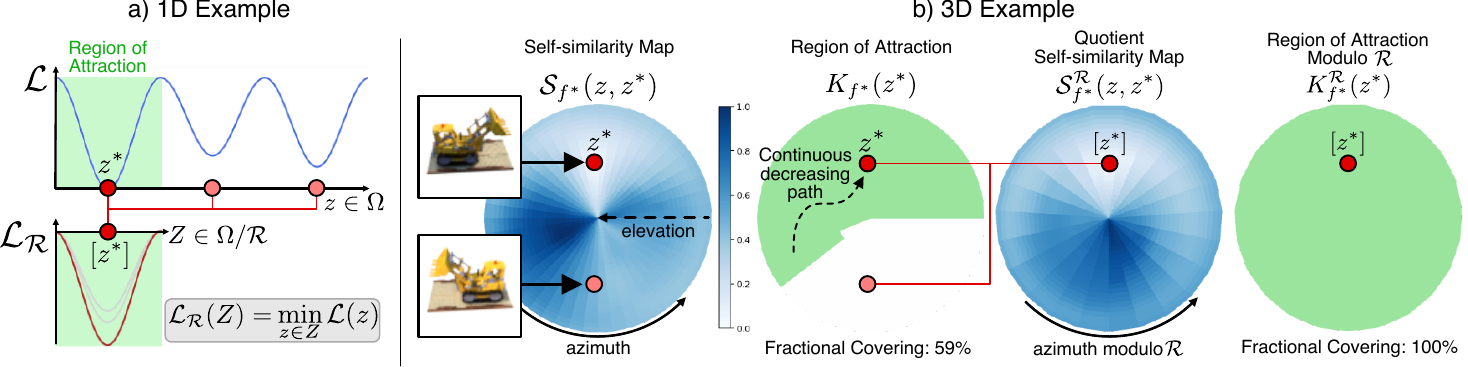}
   \caption{
   \textbf{(a)} One-dimensional quasi-periodic loss $\mathcal{L}$ defined on $\Omega=\sotwo$, minimized by $z^*$, and modulo loss $\mathcal{L}_\relation$ defined on $\Omega/\relation$. The region of attraction is the set of $z$'s that can be connected to $z^*$ with a path where $\mathcal{L}$ always decreases. In this example, the modulo loss is convex for $\relation=\relation_3$.
   \textbf{(b)} Self-similarity map and region of attraction on the Lego scene for all azimuths and elevations, fixed in-plane roll and a fixed reference camera pose $z^*$.
   The self-similarity map shows the photometric distance to the reference image with respect to the rendering pose $z\in\Omega$.  The region of attraction covers all the quotient set $\Omega/\mathcal{R}$ when $\mathcal{R}=\mathcal{R}_2$. See supplements for the self-similarity-map of the Lego scene with a fixed elevation and a variable reference $z^*$.
   }
   \label{fig:method-3d}
\vspace{-3mm}
\end{figure*}

Let $\latentspace$ be the latent space ($\sotwo$ or $\sothree$) and $\nerfcolor$ our rendering model. In both cases, we can provide $\Omega$ with an internal binary operation $\cdot$ and see $(\Omega,\cdot)$ as a group.
In order to focus our analysis on the pose estimation problem, we will make the simplifying assumption that $\shared_\weightsshared$ approximates $\shared^*$, up to a global \textit{alignment} $\alignment\in\latentspace$, \textit{i.e.}
\begin{equation}
    \forall z\in\latentspace,\quad\nerfcolor(.~;\shared^*,\alignment\cdot z) = \nerfcolor(.~;\shared_\weightsshared,z).
\label{eq:approx}
\end{equation}

We define the \textit{self-similarity map} of $\shared^*$ as
\begin{equation}
    \degen_{\shared^*}:\begin{cases}
    \latentspace^2 & \to\mathbb{R}_+\\
    (z,z^*) & \mapsto \norm\big{ \nerfcolor(.~;\shared^*,z) - \nerfcolor(.~;\shared^*,z^*)}_2^2
    \end{cases},
\label{eq:self-similarity}
\end{equation}
Given \eqref{eq:approx}, the loss~\eqref{eqn:generic-loss} can be re-written
\begin{equation}
    \mathcal{L}(\xi) = \sum_{i=1}^n \degen_{\shared^*}(\alignment\cdot\enc_\xi(\image), z^*_i).
\label{eq:loss-poses}
\end{equation}
By the chain rule, the computation of $\nabla_\weightsenc\mathcal{L}$ involves the gradients of $\degen_{\shared^*}(., z^*_i)$. As the loss is minimized by gradient descent, the model can converge to the optimal solution only if it does not need to cross ``bariers'' in the energy landscape to reach a global minimum\def\thefootnote{1}\footnote{for a an infinitely small learning rate}, \textit{i.e.} if there exists a continuous path in $\Omega$ between the initial value $\alignment\cdot h_\weightsenc(\image)$ and $z^*_i$ where $\degen_{\shared^*}(.,z^*_i)$ is strictly decreasing. For $z^*\in\Omega$, we write $\conv_{f^*}(z^*)$ the set of initial values satisfying the above condition and call it the \textit{region of attraction} of $z^*$ for $f^*$.
The difficulty of the pose estimation problem with random initialization can therefore be related to the fractional size of the regions of attraction of $f^*$:
\begin{equation}
    0\leq D(f^*)=\mathbb{E}_{z^*\sim p^*,z\sim p}[z\in\conv_{f^*}(z^*)]\leq 1,
\label{eq:difficulty}
\end{equation}
where $p=p^*$ is the uniform distribution over the latent space $\latentspace$. The closer this value to $1$, the easier pose estimation will be.

\subsection{\LossName: From Latent to Quotient Space}
\label{sec:modulo-loss}

Our core idea is to let the encoder work in a smaller space than the latent space, namely in the quotient set of the latent space for some equivalence relation. Given an equivalence relation $\relation$ (reflexive, transitive and symmetric), we map each latent $\enc_\weightsenc(\image)$ predicted by the encoder to its equivalence class
\begin{equation}
    [\enc_\weightsenc(\image)]_\relation = \{z~|~z\relation\enc_\weightsenc(\image)\}.
\end{equation}
All the latents belonging to this class are passed to the rendering model to generate a set of synthetic observations
\begin{equation}
    \mathcal{C}_i^\relation(\weightsshared,\weightsenc)=\{C(.~;f_\weightsshared,z), z\in[\enc_\weightsenc(\image)]_\relation\}.
\end{equation}
The L2 distances are computed between the ground truth observation and all the generated views.
The function we minimize, called the \textit{\lossname}, is defined as
\begin{equation}
    \mathcal{L}_\relation(\weightsshared,\weightsenc)=\sum_{i=1}^n\text{min}\{\Vert C - I_i \Vert_2^2, C\in\mathcal{C}_i^\relation(\weightsshared,\weightsenc)\}.
\label{eqn:modulo-loss}
\end{equation}
and is a lower bound of~\eqref{eqn:generic-loss} due to the reflexivity of $\relation$.
Given our simplifying assumption~\eqref{eq:approx}, the \lossname can be re-written
\begin{equation}
    \mathcal{L}_\relation(\weightsenc)=\sum_{i=1}^n\degen_{\shared^*}^\relation([\alignment\cdot \enc_\xi(\image)]_\relation,z^*_i)\leq \mathcal{L}(\weightsenc)
\label{eqn:modulo-loss-approx}
\end{equation}
where the \textit{quotient self-similarity map} $\degen_{\shared^*}^\relation$ is defined by
\begin{equation}
    \degen_{\shared^*}^\relation:\begin{cases}
    \latentspace/\relation\times\latentspace & \to\mathbb{R}_+\\
    (Z,z^*) & \mapsto \text{min}\{\degen_{\shared^*}(z^\prime,z^*),z^\prime\in Z\}
    \end{cases}.
\end{equation}
This loss is minimized when $z_i^*\in[\alignment\cdot\enc_\weightsenc(\image)]_\relation$ for all $i$. Given an optimal encoder, $\enc_{\weightsenc^*}(\image)$ is an estimate of $z_i$ \textit{modulo $\relation$} and the true estimate is given by
\begin{equation}
    z_i=\text{argmin}\{\Vert C(.~;f_\weightsshared,z)-\image(.)\Vert_2^2, z\in[\enc_{\weightsenc^*}(\image)]\}.
\label{eqn:predicted-latents}
\end{equation}

As previously, we introduce $\conv^\relation_{f^*}(z^*)$, the region of attraction of $z^*$ for $f^*$ \textit{modulo} $\relation$, as the set of $Z\in\Omega/\relation$ such that there exists a continuous path in $\Omega/\relation$ between $Z$ and $[z^*]_\relation$ where $\degen_{\shared^*}^\relation(.,z^*)$ is strictly decreasing. The difficulty of the pose estimation problem in the quotient space is inversely correlated to the expected fractional covering of  $\conv^\relation_{f^*}(z^*)$,
\begin{equation}
    0\leq D_\relation(f^*)=\mathbb{E}_{z^*\sim p^*,z\sim p}[z\in\conv^\relation_{f^*}(z^*)]\leq 1,
\label{eq:difficulty-rep}
\end{equation}
where $p$ is the uniform distribution over $\latentspace/\relation$ and $p^*$ the uniform distribution over $\latentspace$. Fig.~\ref{fig:method-3d} illustrates the notions of modulo loss, self-similarity map and region of attraction (for a fixed reference $z^*$) in 1D and 3D.

Although $\conv^\relation_{f^*}$ cannot be computed when $\shared^*$ is unknown, the relation $\relation$ can be chosen in a way that matches the general structure of $\degen_{\shared^*}$. For both inverse rendering problems, we introduce an equivalence relation $\relation_{\reporder}$, parameterized by $\reporder\in\mathbb{N}^*$ and defined by
\begin{equation}
    z\relation_{\reporder} z^\prime \quad \text{iff} \quad \exists k\in\mathbb{Z}, z= \frac{2k\pi}{\reporder} + z^\prime.
\end{equation}
$\reporder$ corresponds to the number of distinct elements in $[z]$ and will be called the ``replication order''. In 3D, $\relation_\reporder$ induces a replication of the cameras along the azimuthal dimension and is well-chosen for natural objects, which tend to be pseudo-symmetric by rotation around the vertical axis. For all $\reporder\in\mathbb{N}^*$, a natural homoemorphism $F:\latentspace\to\latentspace/\relation$ is defined by $F(z)=[z/\reporder]$. The quotient space $\latentspace/\relation$ can be seen as ``$\reporder$ times smaller'' than $\latentspace$.

\begin{figure*}[t]
  \centering
  \includegraphics[width=\linewidth]{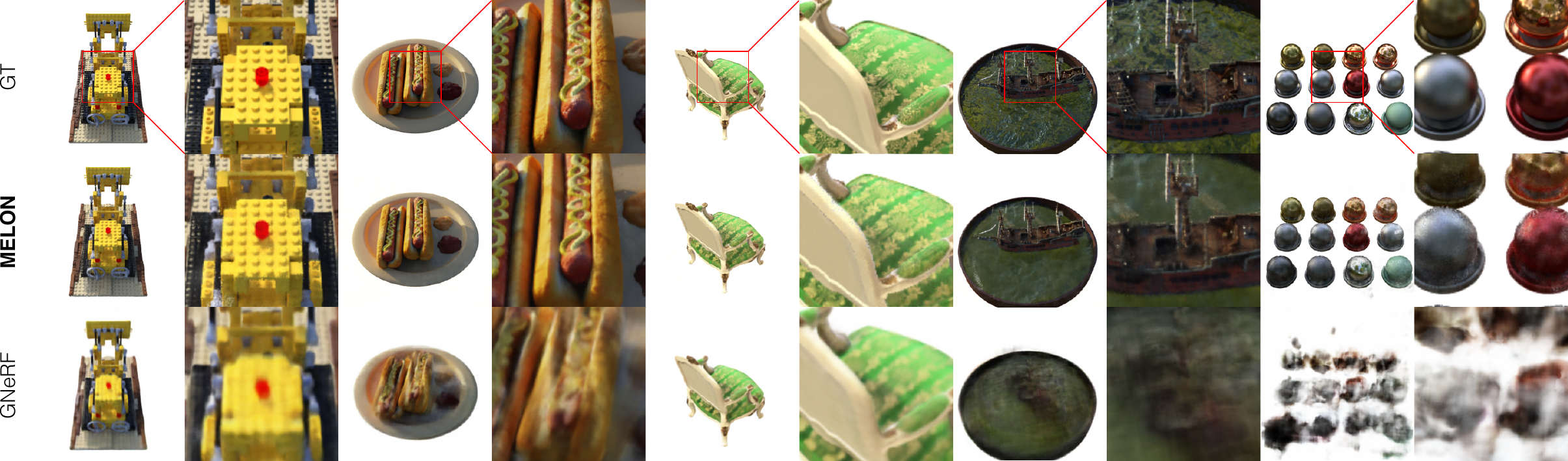}
   \caption{Qualitative comparison between MELON and GNeRF for novel view synthesis (test set) on ``NeRF-Synthetic'' scenes.
   }
   \label{fig:ab-initio-qual}
\end{figure*}

\section{Results}

\subsection{Datasets}
\label{subsec:datasets}

\begin{table*}[t]
  \centering
  \resizebox{\linewidth}{!}{
  \begin{tabular}{lc|cccc|ccc|ccc|ccc}
  \toprule
  & & \multicolumn{4}{c|}{Pose Estimation}
  & \multicolumn{9}{c}{Novel View Synthesis} \\
  
  & & \multicolumn{4}{c|}{Rotation ($^\circ$)~$\downarrow$} 
  & \multicolumn{3}{c|}{PSNR~$\uparrow$} 
  & \multicolumn{3}{c|}{SSIM~$\uparrow$} 
  & \multicolumn{3}{c}{LPIPS~$\downarrow$} \\
  
  \cmidrule(r){3-6} \cmidrule(r){7-9} \cmidrule(r){10-12} \cmidrule(r){13-15}
  
  Scene  & $\reporder$
  & \multicolumn{2}{c}{COLMAP} & GNeRF  & \textbf{\methodname} 
  & GNeRF & \textbf{\methodname} & NeRF
  & GNeRF & \textbf{\methodname} & NeRF
  & GNeRF & \textbf{\methodname} & NeRF \\
  
  \midrule
  
  
  Lego & 2
  & 0.229 & (100) &  2.006  & \textbf{0.061}
  & 24.62 & \textbf{29.07} & 30.44 
  & 0.8756 & \textbf{0.9430} & 0.9526
  & 0.1373 & \textbf{0.0687} & 0.0591 \\
  
  Hotdog & 2
  & 0.328 & (91) & 9.518 & \textbf{0.853}
  & 26.23 & \textbf{31.58} & 35.97 
  & 0.9304 & \textbf{0.9608} & 0.9782 
  & 0.1284 & \textbf{0.0669} & 0.0384  \\
  
  Chair & 2
  & 9.711 & (100) & 1.181 & \textbf{0.124}
  & 29.45 & \textbf{29.72} & 32.45
  & 0.9346 & \textbf{0.9491} & 0.9673
  & 0.0988 & \textbf{0.0708} & 0.0558 \\
  
  Drums    & 2
  & 0.467 & (21) & 1.602 & \textbf{0.048}
  & 22.58 & \textbf{24.38} & 24.43
  & 0.8893 & \textbf{0.9091} & 0.9120
  & 0.1577 & \textbf{0.1169} & 0.1095 \\
  
  Mic & 2
  & 0.860 & (15) & 1.535 & \textbf{0.057}
  & 28.32 & \textbf{28.66} & 31.83 
  & 0.9527 & \textbf{0.9582} & 0.9705
  & 0.0765 & \textbf{0.0620} & 0.0465 \\
  
  Ship  & 2
  & 0.292 & (10) & 27.84 & \textbf{0.113}
  & 15.72 & \textbf{26.87} & 27.71
  & 0.7704 & \textbf{0.8390} & 0.8553 
  & 0.4068 & \textbf{0.1847} & 0.1719  \\
  
  Materials  & 4
  &  1.053 & (17) &  38.43 & \textbf{5.607}
  & 13.13 & \textbf{23.43} & 29.57 
  & 0.7538 & \textbf{0.8914} & 0.9540 
  & 0.4031 & \textbf{0.1620} & 0.0725 \\
  
  Ficus   & 2
  & \multicolumn{2}{c}{fails} & 45.88 & \textbf{8.977}
  & 17.85 & \textbf{22.85} & 25.63 
  & 0.8493 & \textbf{0.9132} & 0.9366  
  & 0.2594 & \textbf{0.1321} & 0.0892  \\

 
  
  
  \bottomrule 
  \end{tabular}
  } 
  
  \caption{Pose estimation (training set) and novel view synthesis (test set) quality for competing methods on ``NeRF-Synthetic'' scenes.
  $\reporder$ is the replication order used with \methodname.
  ``NeRF'' is our implementation with ground truth cameras for baseline reference.
  GNeRF and \methodname predict the elevation only within the respective range of the scene (see supp.)
  We report the best of five runs for GNeRF and MELON.
  COLMAP column left: mean error on solved poses, right: number of solved poses.
  We penalize unsolved COLMAP poses with the mean of a random guess when bolding for fair comparison to GNeRF and \methodname, which always predict pose.
  }
  \label{tab:ab-initio_comparison}
\vspace{-3mm}
\end{table*}


\textbf{1D Dataset}
We generate random functions $f^*$ that map $\sotwo$ to $\mathbb{R}$ by randomly drawing their $512$ first Fourier coefficients $c_k$ following a centered Gaussian distribution. The expected magnitudes of these Fourier coefficients is $\exp(-|k|/5)$ and we make the functions quasi-periodic by setting $c_{\pm 2}=100$. We generate ten one-dimensional functions containing $256$ crops of $65$ uniformly spaced samples of $f^*$. We show an example of ground truth function and its self-similarity map in the supplements.

\textbf{RGB-MELON} 
We build a pathological dataset containing centered views of 3D spheres on which we map quasi-symmetric textures. We use spherical harmonics to generate red-green textures that are invariant by translation of $2\pi/K$ along the azimuthal direction, with a tunable \textit{symmetry order} parameter $K\in\{1,2,3,4\}$. In order to break the perfect symmetry of these scenes, we add three red/green/blue squares along the azimuthal direction. We generate $116$ views from a fixed and known distance, with all the cameras located in the equator plane. We hold out $16$ test images in each dataset. Sample views of these datasets are provided in the supplements.

\textbf{``NeRF-Synthetic'' Dataset} We use the eight object-centric scenes provided by \cite{Mildenhall20eccv_nerf}, which consist of $300$ rendered images (downscaled to $400\times 400$ pixels) and ground truth camera poses for each scene. All the methods evaluated on these datasets assume the object-to-camera distance and the ``up'' direction to be known (fixed roll $\alpha$). We hold out the $200$ ``test'' views, using them only for evaluation.

\textbf{Real-World Datasets} We evaluate our method using three real-world scenes: the Gold Cape~\cite{goldcape} and the Ethiopian Head~\cite{ethiopianhead} from the British Museum's photogrammetry dataset, and a Toytruck from the Common Object in 3D (CO3D) dataset~\cite{reizenstein2021common}. These scenes are approximately object-centered and were recorded with fixed illumination conditions. We consider the published camera poses as ground truth. We use ground truth masks to remove the background before computing the photometric loss.

\subsection{1D Inverse Rendering}
\label{subsec:1dinverse}
In our 1D study we map crops to predicted angles with an encoder, trivially "render" with the identity following \eqref{eqn:one-d-forward-model}, use the modulo loss and reconstruct the function $f_\weightsshared$ with a neural network (positional encoding \cite{Mildenhall20eccv_nerf} and four fully-connected layers of size 64 with ReLU activation functions). We perform an ablation study and separately (1) replace the neural representation $f_\weightsshared$ with an explicit representation, (2) replace the modulo loss with an L2 loss and (3) replace the encoder with a set of independently optimized angles. We use the Adam optimizer \cite{kingma2014adam} with a learning rate of $10^{-4}$ and $256$ crops per batch. We align the predicted angles (as defined by~\eqref{eqn:predicted-latents}) and the predicted function to the ground truth before computing mean square errors. As shown in Fig.~\ref{fig:torus-results}, the neural representation, the modulo loss and the encoder are all necessary for the method to work, in this order of importance. We hypothesize that explicit grids are not well-suited for this task because they do not have any inductive bias encouraging smoothness. Since the derivative of the loss w.r.t. $\theta_i$ involves $f_\weightsshared^\prime$, gradient-descent gets stuck at an early stage (visual reconstructions in the supplements). The modulo loss mods out the latent space, making it more convex. The encoder leverages the fact that similar images should be mapped to similar poses.

\begin{figure}
    \centering
    \includegraphics[width=\linewidth]{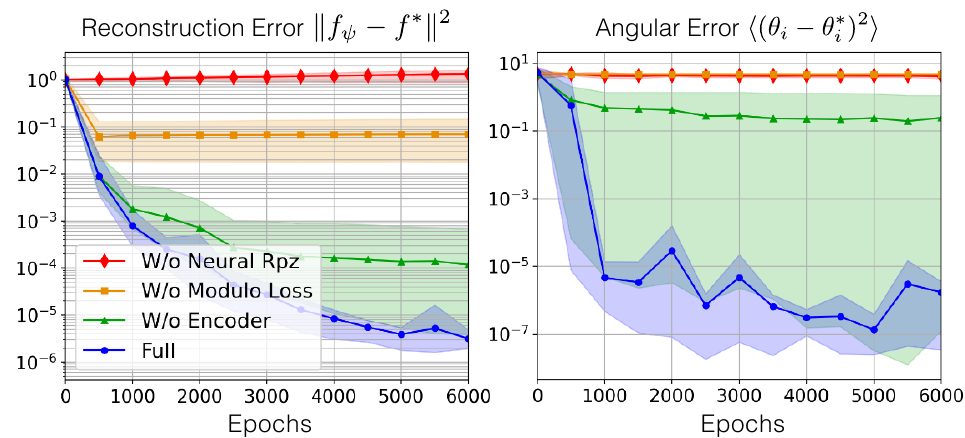}
    \caption{Angular and reconstruction error for the 1D datasets. We indicate the mean, min and max errors obtained throughout $10$ experiments. The explicit representation gets stuck at an early stage. The modulo loss helps the model avoiding local minima and the encoder regularizes the angular prediction.
    }
    \label{fig:torus-results}
\end{figure}

\subsection{NeRF with Unposed Images}

Using the ``NeRF-Synthetic'' dataset, we compare \methodname to COLMAP~\cite{Schonberger2016sfmrevisited} and GNeRF~\cite{Meng21iccv_GNeRF} for pose estimation and novel view synthesis. In Table~\ref{tab:ab-initio_comparison}, we quantify the accuracy of pose estimation by reporting the mean angular error of the predicted view directions on the training set. We evaluate novel view synthesis quality by computing the Peak Signal Noise Ratio (PSNR), Structural Similarity Index Measure (SSIM)~\cite{ssim} and Learned Perceptual Image Patch Similarity (LPIPS)~\cite{lpips} against all 200 test views. The pose accuracy is reported after applying a global alignment matrix minimizing the mean angular error over $\sothree$. A qualitative comparison is shown on Fig.~\ref{fig:ab-initio-qual}. \methodname outperforms GNeRF on novel view synthesis, thanks to more accurate estimation of the camera poses. COLMAP fails on scenes where it does not find enough corresponding points (``ficus'') while a small number of outliers significantly increases the mean angular error on ``chair''. For fair comparison to GNeRF and MELON, we penalize COLMAP's unsolved poses with a random guess (see corrected pose accuracy in the supplements).

\begin{figure}
  \centering
  \includegraphics[width=\linewidth]{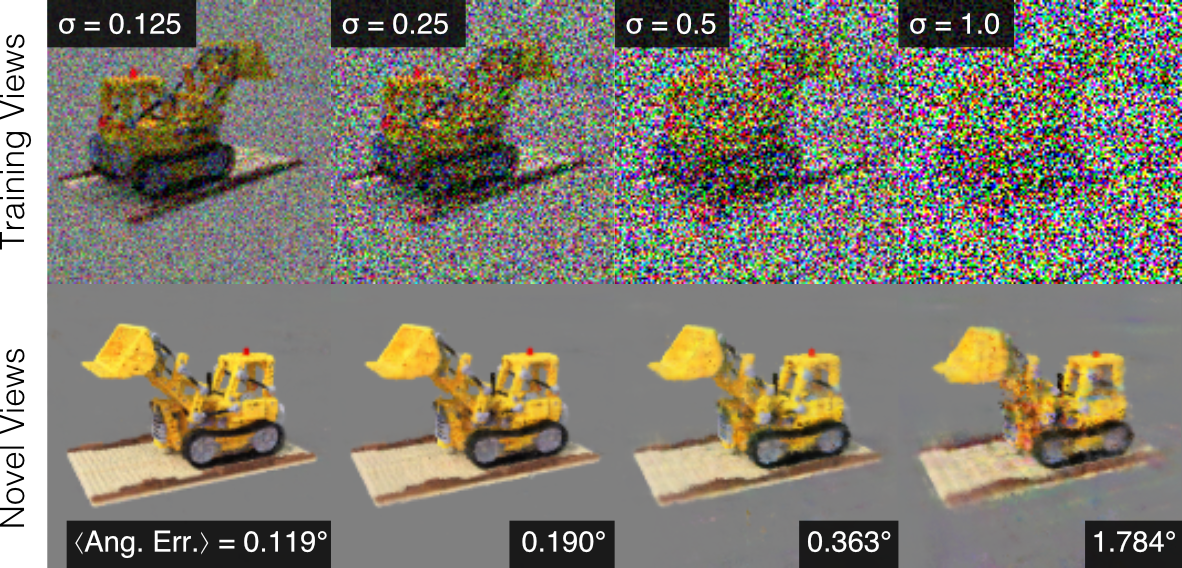}
   \caption{Novel view synthesis from noisy unposed images ($128\times128$). Mean angular error on the training set indicated in bottom right corners.}
   \label{fig:noise-sweep}
\vspace{-3mm}
\end{figure}

\subsection{Noise and Number of Views}

We illustrate the sensitivity of \methodname to noise in Fig.~\ref{fig:noise-sweep}. We add white Gaussian noise of variance $\sigma^2$ to the training images of the ``lego'' dataset. For this experiment, we encourage view-consistency using a view-independent neural field.
\methodname accurately estimates the poses on the training set and generates noise-free novel views. In this context, our method can be seen as a ``3D-consistent'' unposed image denoiser.
A comparison to COLMAP is provided in the supplements.

Unlike adversarial approaches, \methodname does not need to train a model to discriminate real from synthetic images, and can therefore perform \textit{ab initio} inverse rendering with datasets containing few images. We artificially reduce the size of the training set in the ``lego'' scene and report the results of \methodname with a replication order $\reporder=2$ in Table~\ref{tab:num_views_sweep}. We compare our method to the adversarial approach GNeRF, and to COLMAP. Novel views are shown in the supplements.

COLMAP does not find enough corresponding points when given 16 or less views. In comparison, \methodname accurately estimates the poses from only six views while GNeRF's accuracy significantly degrades when decreasing the number of given views. We hypothesize that supervising the field with slightly wrong cameras at an early stage of training can act as a regularizer and could explain why \methodname sometimes gets a marginally higher PSNR than NeRF with ground truth poses.

\begin{table}[t]
  \centering
  \resizebox{\linewidth}{!}{
  \begin{tabular}{l|cccc|ccc}
  \toprule
  
  & \multicolumn{4}{c|}{Rotation ($^\circ$) $\downarrow$} 
  & \multicolumn{3}{c}{PSNR $\uparrow$} \\

  \cmidrule(r){2-5} \cmidrule(r){6-8} 

  \# Views
  & \multicolumn{2}{c}{COLMAP} & GNeRF &\textbf{\methodname} 
  & GNeRF &  \textbf{\methodname} & NeRF  \\

  \midrule
  
  4 &
  \multicolumn{2}{c}{fails} & 45.8 & \textbf{1.21} &
  12.59 & \textbf{12.83} & 18.47 \\

  6 &
  \multicolumn{2}{c}{fails} & 35.4 & \textbf{0.509} &
  13.20 & \textbf{18.95} & 19.75 \\
  
  8 &
  \multicolumn{2}{c}{fails} & 20.0 & \textbf{0.517} &
  14.13 & \textbf{22.57} & 24.08 \\

  12 &
  \multicolumn{2}{c}{fails} & 17.9 & \textbf{0.148} &
  15.02 & \textbf{26.40} & 26.29 \\

  16 &
  \multicolumn{2}{c}{fails} & 11.7 & \textbf{0.135} &
  16.79 & \textbf{27.33} & 27.20 \\

  32 &
  41.2 & (20) &  12.3 & \textbf{0.122} &
  18.43 & \textbf{29.01} & 29.07 \\

  64 &
  35.3 & (64) & 4.77 & \textbf{0.101} &
  23.74 & \textbf{30.80} & 30.31 \\

  100 &
  7.66 & (100) & 0.901 & \textbf{0.095} &
  29.24 & \textbf{31.15 } & 30.72 \\

  \bottomrule 
  \end{tabular}
  } 

  \caption{Ab-initio reconstruction with a variable number of views on the ``lego'' scene (128$\times$128). ``NeRF'' is our implementation with ground truth cameras.
  We report the best of three runs for GNeRF and MELON.
  COLMAP numbers reported as in Table~\ref{tab:ab-initio_comparison}.
  }
  \label{tab:num_views_sweep}
\vspace{-3mm}
\end{table}


\subsection{Replication Order}

\begin{table*}[t]
  \centering
  \resizebox{\linewidth}{!}{
  \begin{tabular}{c|ccc|ccc|ccc|ccc|c}
  \toprule
  & \multicolumn{12}{c|}{\textbf{\methodname}}
  & COLMAP \\
  
  & \multicolumn{3}{c|}{$\reporder=1$ (23 ms/it.)}
  & \multicolumn{3}{c|}{$\reporder=2$ (34 ms/it.)}
  & \multicolumn{3}{c|}{$\reporder=3$ (43 ms/it.)}
  & \multicolumn{3}{c|}{$\reporder=4$ (52 ms/it.)}
  &  \\
  
  \cmidrule(r){2-13}
  
  Sym. Order
  & $D(f^*)$ & PSNR & Rot. ($^\circ$) 
  & $D_\relation(f^*)$ & PSNR & Rot. ($^\circ$) 
  & $D_\relation(f^*)$ & PSNR & Rot. ($^\circ$) 
  & $D_\relation(f^*)$ & PSNR & Rot. ($^\circ$)
  & Rot. ($^\circ$) \\
  \midrule
  1 & 
  \cellcolor{Green} 0.65 & \cellcolor{Green} 36$\pm$1 & \cellcolor{Green} 2.7$\pm$0.5 & 
  \cellcolor{Green} 0.83 & \cellcolor{Green} 36$\pm$1 & \cellcolor{Green} 2.6$\pm$0.5 & 
  \cellcolor{Green} 0.86 & \cellcolor{Green} 37$\pm$1 & \cellcolor{Green} 2.5$\pm$0.5 & 
  \cellcolor{Green} 1.00 & \cellcolor{Green} 37$\pm$1 & \cellcolor{Green} 2.5$\pm$0.5 &
  fails \\
  
  2 & 
  \cellcolor{Red} 0.52 & \cellcolor{Red} 25$\pm$2 & \cellcolor{Red} 70$\pm$5 & 
  \cellcolor{Green} 0.98 & \cellcolor{Green} 38.6$\pm$0.5 & \cellcolor{Green} 0.7$\pm$0.1 & 
  \cellcolor{Red} 0.55 & \cellcolor{Red} 19$\pm$1 & \cellcolor{Red} 70$\pm$5 & 
  \cellcolor{Green} 1.00 & \cellcolor{Green} 38.7$\pm$0.5 & \cellcolor{Green} 0.7$\pm$0.1 &
  fails \\
  
  3 & 
  \cellcolor{Red} 0.28 & \cellcolor{Red} 16$\pm$1 & \cellcolor{Red} 72$\pm$5 & 
  \cellcolor{Red} 0.40 & \cellcolor{Red} 21$\pm$2 & \cellcolor{Red} 40$\pm$5 & 
  \cellcolor{Green} 0.65 & \cellcolor{Green} 39.5$\pm$0.5 & \cellcolor{Green} 0.6$\pm$0.1 & 
  \cellcolor{Red} 0.57 & \cellcolor{Red} 29$\pm$5 & \cellcolor{Red} 25$\pm$11 &
  fails \\
  
  4 & 
  \cellcolor{Red} 0.26 & \cellcolor{Red} 22$\pm$1 & \cellcolor{Red} 75$\pm$5 & 
  \cellcolor{Red} 0.54 & \cellcolor{Red} 26$\pm$1 & \cellcolor{Red} 47$\pm$5 & 
  \cellcolor{Red} 0.35 & \cellcolor{Red} 19$\pm$1 & \cellcolor{Red} 70$\pm$5 & 
  \cellcolor{Green} 1.00 & \cellcolor{Green} 35$\pm$1 & \cellcolor{Green} 3.0$\pm$0.5 &
  fails \\
  
  \bottomrule 
  \end{tabular}
  } 
  
  \caption{The fractional size of the regions of attractions $D_\relation(f^*)$ gets closer to $1$ when the replication order $\reporder$ is a multiple of the symmetry order $K$ of the sphere. We report the accuracy of novel view synthesis and pose estimation (mean $\pm$ std. over 5 experiments). COLMAP fails due to the repeating patterns in the synthesized textures. Red cells indicate when the mean angular error is above $20^\circ$. The compute time of one gradient-step increases linearly with $N$.}
  \label{tab:sph-results}
\vspace{-3mm}
\end{table*}

We run \methodname with a replication order $\reporder\in\{1,2,3,4\}$ to reconstruct the RGB-MELON scenes. In Table~\ref{tab:sph-results}, we report, for each value of $\reporder$ and each symmetry order $K$, the fractional size $D_\relation(f^*)$ of the regions of attraction for $\relation=\relation_\reporder$. This number is closer to $1$ when the replication order ``matches'' the symmetry order of the dataset, \textit{i.e.} when $N$ is a multiple of $K$.

We report the quality of novel view synthesis on the test set and pose estimation on the training set. COLMAP fails at estimating the poses due to the presence of repeating patterns and the absence of sharp corners. In comparison, our method converges when the replication order matches the symmetries of the texture. As $\reporder$ increases, our method can cope with a broader range of datasets at the cost of more computation.

\subsection{Towards NeRF in the Wild}

\methodname assumes that all the cameras point towards the center of the scene from a known distance. In a real capture setup, the object-to-camera distance, $r$, is unknown and the cameras do not all point towards a unique point. We parameterize each camera with a camera-to-world rotation matrix $R_i\in\sothree$ and a three-dimensional vector $T_i\in\mathbb{R}^3$ that specifies the location of the origin in the camera frame. This vector is constant and equal to $(0, 0, r)$ in an ideal object-centered setup.

In Fig.~\ref{fig:nerf_in_the_wild}, we compare our results to those obtained with GNeRF~\cite{Meng21iccv_GNeRF} and SAMURAI~\cite{Boss2022samurai}, for the estimation of the rotation matrix $R_i$ only. For all methods, we use the published values for $T_i$ and parameterize the elevations in $[0^\circ,90^\circ]$. Since SAMURAI requires the user to manually initialize the camera view directions to one of 26 possible directions, we show the results obtained with a fixed initialization at the North pole for fair comparison. Quantitative results and additional comparison to a manual initialization of SAMURAI are shown in the supplements.

When both the object-to-camera distances and the in-plane camera translations are unknown, the latent space becomes $\text{SE}(3)$ and contains twice as many degrees of freedom as $\text{SO}(3)$. Although it can accurately estimate the rotation matrix $R_i$, we show in Fig.~\ref{fig:limitations} that \methodname fails at jointly inferring $R_i$ and $T_i$. In this setup, SAMURAI is the only effective reconstruction method but requires a rough initialization of the camera view directions (see supplements).

\begin{figure}
  \centering
  \includegraphics[width=\linewidth]{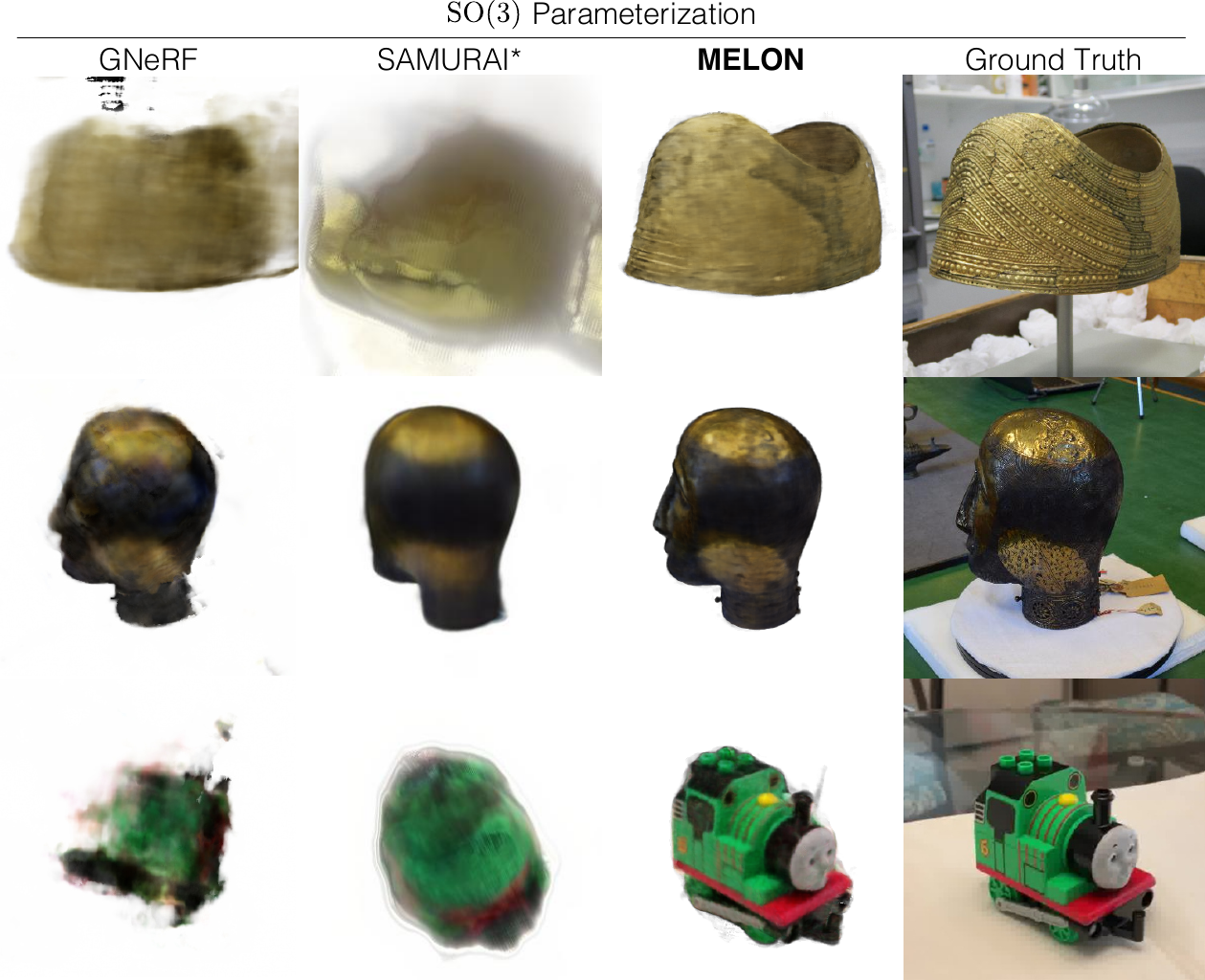}
   \caption{Ab initio reconstruction on real datasets. All the methods use ground truth values for the object-to-camera distances and in-plane camera translations. SAMURAI* uses a fixed initialization of the poses at the North pole.
   }
   \label{fig:nerf_in_the_wild}
\vspace{-3mm}
\end{figure}
\begin{figure}
  \centering
  \includegraphics[width=\linewidth]{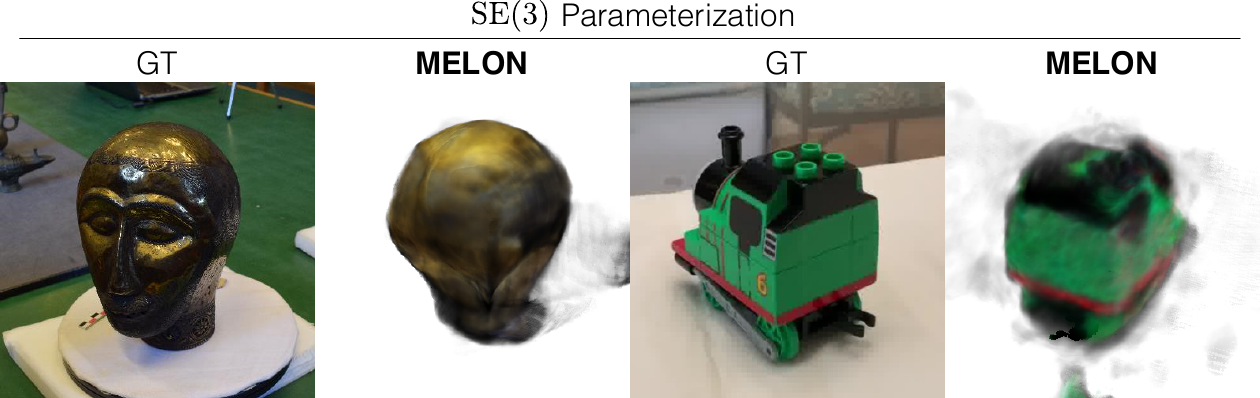}
   \caption{Limitation: \methodname fails when the object-to-camera distance and in-plane camera translations are unknown.}
   \label{fig:limitations}
\vspace{-3mm}
\end{figure}

\section{Discussion}

\methodname solves the problem of inverse rendering from unposed images using amortized inference and a novel loss function.
Using an equivalence relation that matches the distribution of local minima in a given object's self-similarity map, the \lossname reduces the camera space to a smaller space (its quotient set), in which gradient descent is more likely to converge.
We demonstrated that \methodname could perform inverse rendering on a variety of synthetic and real unposed datasets with state-of-the-art accuracy and that \methodname could cope with small or noisy datasets.

Characterizing the full loss landscape of 3D inverse rendering with unknown poses remains an open research question.
Our theoretical analysis focused on the pose estimation problem only, making the assumption that the reconstructed function $f$ was close to $f^*$, up to a global alignment.
Although the loss~\eqref{eqn:generic-loss} simplified to~\eqref{eq:loss-poses}, no inequality generally holds between the two. The full inverse rendering problem contains many more unknown parameters and ambiguities than pose estimation alone.
Furthermore, the assumption of a perfectly object-centered setup is too restrictive for reconstructing real world datasets and the prediction of camera extrinsics in $\text{SE}(3)$ remains an open challenge.

We demonstrated the possibility of optimizing a neural radiance field from unposed images. However, other volumetric representations are conceivable, such as tensorial products~\cite{Chan2021eg3d, Chen2022tensorf} hash tables~\cite{Muller2022ingp}, or explicit representations~\cite{fridovich2022plenoxels}. Exploring the strengths of other encodings within the context of our method is a promising future direction.

As real objects often contain rotational symmetries around the vertical axis, we focused on the use of equivalence classes that keep the elevation and roll of the cameras fixed. An interesting extension of this work could be to broaden the scope of equivalence relations used to mod out the latent space $\sothree$. 

\section*{Acknowledgements}

We thank Matthew Brown, Ricardo Martin-Brualla and Fr{\'e}d{\'e}ric Poitevin and Rick Szeliski for their helpful technical insights and feedback.

We acknowledge the use of the computational resources at the SLAC Shared
Scientific Data Facility (SDF).

\newpage

{\small
\bibliographystyle{ieeenat_fullname}
\bibliography{nerf,non-nerf}
}

\clearpage

}

\end{document}